\documentclass[10pt,twocolumn,letterpaper]{article}
\pdfoutput=1
\usepackage{iccv}
\usepackage{times}
\usepackage{epsfig}
\usepackage{graphicx}
\usepackage{amsmath}
\usepackage{amssymb}
\usepackage{bm}
\usepackage{multirow}

\usepackage[breaklinks=true, colorlinks, bookmarks=false]{hyperref}

\iccvfinalcopy 

\def\iccvPaperID{1323} 

\ificcvfinal\fi
\begin{document}

\title{GridDehazeNet: Attention-Based Multi-Scale Network for Image Dehazing}


\newcommand*\samethanks[1][\value{footnote}]{\footnotemark[#1]}

\author{Xiaohong Liu\thanks{Authors contributed equally.} \quad Yongrui Ma\samethanks[1] \quad Zhihao Shi\quad Jun Chen \\McMaster University\\
	{\tt\small \{liux173, may85, shiz31, chenjun\}@mcmaster.ca}
}

\maketitle
\ificcvfinal\thispagestyle{empty}\fi

\begin{abstract}
	We propose an end-to-end trainable Convolutional Neural Network (CNN), named GridDehazeNet, for single image dehazing. The GridDehazeNet consists of three modules: pre-processing, backbone, and post-processing. The trainable pre-processing module can generate learned inputs with better diversity and more pertinent features as compared to those derived inputs produced by hand-selected pre-processing methods. The backbone module implements a novel attention-based multi-scale estimation on a grid network, which can effectively alleviate the bottleneck issue often encountered in the conventional multi-scale approach. The post-processing module helps to reduce the artifacts in the final output. Experimental results indicate that the  GridDehazeNet outperforms the state-of-the-arts on both synthetic and real-world images. The proposed hazing method does not rely on the atmosphere scattering model, and we provide an explanation as to why it is not necessarily beneficial to take advantage of the dimension reduction offered by the atmosphere scattering model for image dehazing, even if only the dehazing results on synthetic images are concerned. Project website: \url{https://proteus1991.github.io/GridDehazeNet/}.
	
\end{abstract}

\section{Introduction}

\begin{figure}[t]
	\centering
	
	\begin{minipage}[h]{0.49\linewidth}
		\centering
		\includegraphics[width=\linewidth]{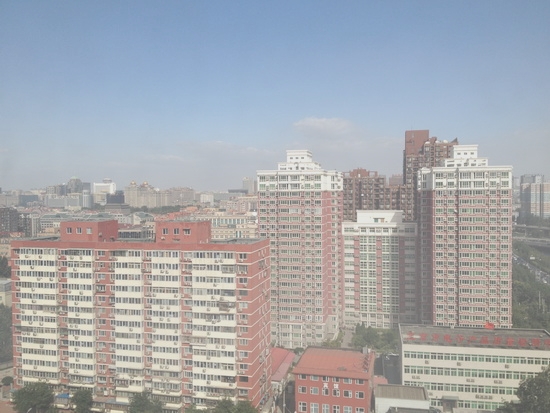}
		\scriptsize{(a) Hazy Image}
	\end{minipage}
	\begin{minipage}[h]{0.49\linewidth}
		\centering
		\includegraphics[width=\linewidth]{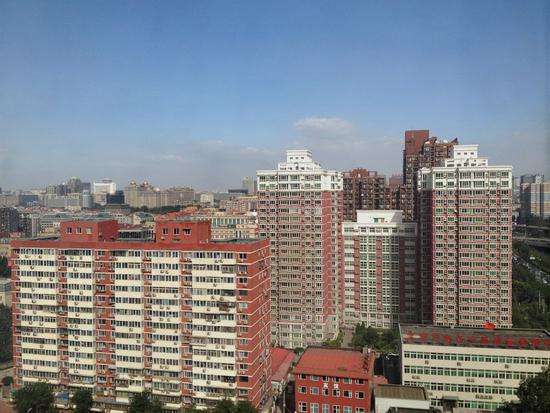}
		\scriptsize{(b) Our dehazed Image}
	\end{minipage}
	\vspace{0.3em}
	\caption{An example of image dehazing.}
	\label{fig:first_page}
	\vspace{-1.5em}
\end{figure}

The image dehazing problem has received significant attention in the computer vision community over the past two decades. Image dahazing aims to recover the clear version of a hazy image (see Fig.~\ref{fig:first_page}). It helps mitigate the impact of image distortion induced by the environmental conditions on various visual analysis tasks, which is essential for the development of robust intelligent surveillance systems. 

The atmosphere scattering model~\cite{scatteringfn01, scatteringfn02, scatteringfn03} provides a simple approximation of the haze effect. Specifically, it assumes that
\begin{align}
I_i(x)=J_i(x)t(x)+A(1-t(x)),\quad i=1,2,3,\label{eq:asm}
\end{align}
where $I_i(x)$ ($J_i(x)$) is the intensity of the $i$th color channel  of pixel $x$ in the hazy (clear) image, $t(x)$ is the transmission map, and $A$ is the global atmospheric light intensity; moreover, we have $t(x)=e^{-\beta d(x)}$ with $\beta$ and $d(x)$ being the atmosphere scattering parameter and the scene depth, respectively. 
This model indicates that image dehazing is in general an underdetermined problem without the knowledge of $A$ and $t(x)$.

As a canonical example of image restoration, the dehazing problem can be tackled using a variety of techniques that are generic in nature. Moreover, many misconceptions and difficulties encountered in image dehazing manifest in other restoration problems as well.
Therefore, it is instructive to examine the relevant issues in a broader context, three of which are highlighted below.

	1. Role of physical model: Many data-driven approaches to image restoration require synthetic datasets for training. To create such datasets, it is necessary to have a physical model of the relevant image degradation process ({\textit{e}.\textit{g}.,  the atmosphere scattering model for the haze effect).  A natural question arises whether the design of the image restoration algorithm itself should rely on this physical model. Apparently a model-dependent algorithm may suffer inherent performance loss on real-world images due to model mismatch. However, it is often taken for granted that such an algorithm must have advantages on synthetic images created using the same physical model.
	
	2. Selection of pre-processing method: Pre-processing is widely used in image preparation to facilitate follow-up operations~\cite{tong2017image, iebmgated01}. It can also be used to generate several variants of the given image, providing a certain form of diversity that can be harnessed via proper fusion. However, the pre-processing methods are often selected based on heuristics, thus are not necessarily best suited to the problem under consideration.

	3. Bottleneck of multi-scale estimation: 	
	Image restoration requires an explicit/implicit knowledge of the statistical relationship between the distorted image and the original clear version. The statistical model needed to capture this relationship often has a huge number of parameters, comparable or even more than the available training data. As such, directly estimating these parameters based on the training data is often  unreliable. Multi-scale estimation~\cite{shen2018deep, chen2018learning} tackles this problem by i) approximating the high-dimensional statistical model by a low-dimensional one, ii) estimating the parameters of the low-dimensional model based on the training data, iii) parameterizing the neighborhood of the estimated low-dimensional model,  performing a refined estimation, and repeating this procedure if needed. It is clear that the estimation accuracy on one scale will affect that on the next scale. Since multi-scale estimation is commonly done in a successive manner, its performance is often limited by a certain bottleneck.


The main contribution of this work is an end-to-end trainable CNN, named GridDehazeNet, for single image dehazing. This network can be viewed as a product of our attempt to address the aforementioned generic issues in image restoration. Firstly, the proposed GridDehazeNet does not rely on the atmosphere scattering model in Eq.~(\ref{eq:asm}) for haze removal, yet is capable of outperforming the existing model-dependent dehazing methods even on synthetic images; a possible explanation, together with some supporting experimental results, is provided for this puzzling phenomenon. Secondly, the pre-processing module of GridDehazeNet is fully trainable; the learned pre-processor can offer more flexible and pertinent image enhancement as compared to hand-selected pre-processing methods. Lastly, the implementation of attention-based multi-scale estimation on a grid network allows efficient information exchange across different scales and alleviate the bottleneck issue. It will be shown that the proposed dehazing method achieves superior performance in comparison with the state-of-the-arts.

\section{Related Work}
Early works on image dehazing either require multiple images of the same scene taken under different conditions~\cite{multiimagepolar01, multiimagepolar02, scatteringfn02, multiimageweather02, multiimageweather03} or side information acquired from other sources~\cite{multiimagedepth02,multiimagedepth01}.

Single image dehazing with no side information is considerably more difficult. Many methods have been proposed to address this challenge. A conventional strategy is to estimate the transmission map $t(x)$ and the global atmospheric light intensity $A$ (or their variants) based on certain assumptions or priors then invert Eq.~(\ref{eq:asm}) to obtain the dehazed image. Representative works along this line of research include~\cite{iebmcontrast01,iebmalbedo01,iebmdcp01,iebmrf01,iebmlinear01}. Specifically, \cite{iebmcontrast01} proposes a local contrast maximization method for dehazing based on the observation that clear images tend to have higher contrast as compared to their hazy counterparts; in~\cite{iebmalbedo01} haze removal is realized via the analysis of albedo under the assumption that the transmission map and surface shading are locally uncorrelated; the dehazing method introduced in~\cite{iebmdcp01} makes use of the Dark Channel Prior (DCP), which asserts that pixels in non-haze patches have low intensity in at least one color channel; \cite{iebmrf01} suggests a machine learning approach that exploits four haze-related features using a random forest regressor; the color attenuation prior is adopted in~\cite{iebmlinear01} for the development of  a supervised learning method for image dehazing. Although these methods have enjoyed varying degrees of success, their performances are inherently limited by the accuracy of the adopted assumptions/priors with respect to the target scenes.

 With the advance in deep learning technologies and the availability of large synthetic datasets~\cite{iebmrf01}, recent years have witnessed the increasing popularity of data-driven methods for image dehazing. These methods largely follow the conventional strategy mentioned above but with reduced reliance on hand-crafted priors. For example, the dehazing method, DehazeNet, proposed in~\cite{iebmdehazenet01} uses a three-layer CNN to directly estimate the transmission map from the given hazy image; \cite{iebmmscnn01} employs a Multi-Scale CNN (MSCNN) that is able to perform refined transmission estimation.
 
  The AOD-Net~\cite{iebmaod01} represents a departure from the conventional strategy. Specifically, a reformulation of Eq.~(\ref{eq:asm}) is introduced in~\cite{iebmaod01} to bypass the estimation of the transmission map and the atmospheric light intensity. 
 A close inspection reveals that this reformulation in fact renders the atmosphere scattering model completely superfluous (though this point is not recognized in~\cite{iebmaod01}). \cite{iebmgated01} goes one step further by explicitly abandoning  the atmosphere scattering model in algorithm design. The Gated Fusion Network (GFN) proposed in~\cite{iebmgated01} leverages hand-selected pre-processing methods and multi-scale estimation, which are generic in nature and are subject to improvement.

\section{GridDehazeNet}

The proposed GridDehazeNet is an end-to-end trainable network with three important features.

	1. No reliance on the atmosphere scattering model: Among the aforementioned single image dehazing methods, only AOD-Net and GFN do not rely on the atmosphere scattering model.
	However, no convincing reason has been provided why there is any advantage in ignoring this model, as far as the dehazing results on synthetic images are concerned.
	The argument put forward in~\cite{iebmgated01} is that estimating $t(x)$ from a hazy image is an ill-posed problem. Nevertheless, this is puzzling since estimating $t(x)$ (which is color-channel-independent) is presumably easier than $J_i(x)$, $i=1,2,3$. 
	In Fig.~\ref{fig:losssurface} we offer a possible explanation why it could be problematic if one blindly uses the fact that $t(x)$ is color-channel-independent to narrow down the search space and why it might be potentially advantageous to relax this constraint in the search of the optimal $t(x)$. However, with this relaxation, the atmosphere scattering model offers no dimension reduction in the estimation procedure. 
	More fundamentally, it is known that the loss surface of a CNN is generally well-behaved in the sense that the local minima are often almost as good as the global minimum~\cite{choromanska2015loss,draxler2018essentially,nguyen2018loss}. On the other hand, by incorporating the atmosphere scattering model into a CNN, one basically introduces a nonlinear component that is heterogeneous in nature from the rest of the network, which may create an undesirable loss surface. To support this explanation, we provide some experimental results in Section \ref{sec:asm}.
	
	\begin{figure}[h]
		\centering
		\begin{minipage}[h]{0.45\linewidth}
			\centering
			\includegraphics[width=\linewidth]{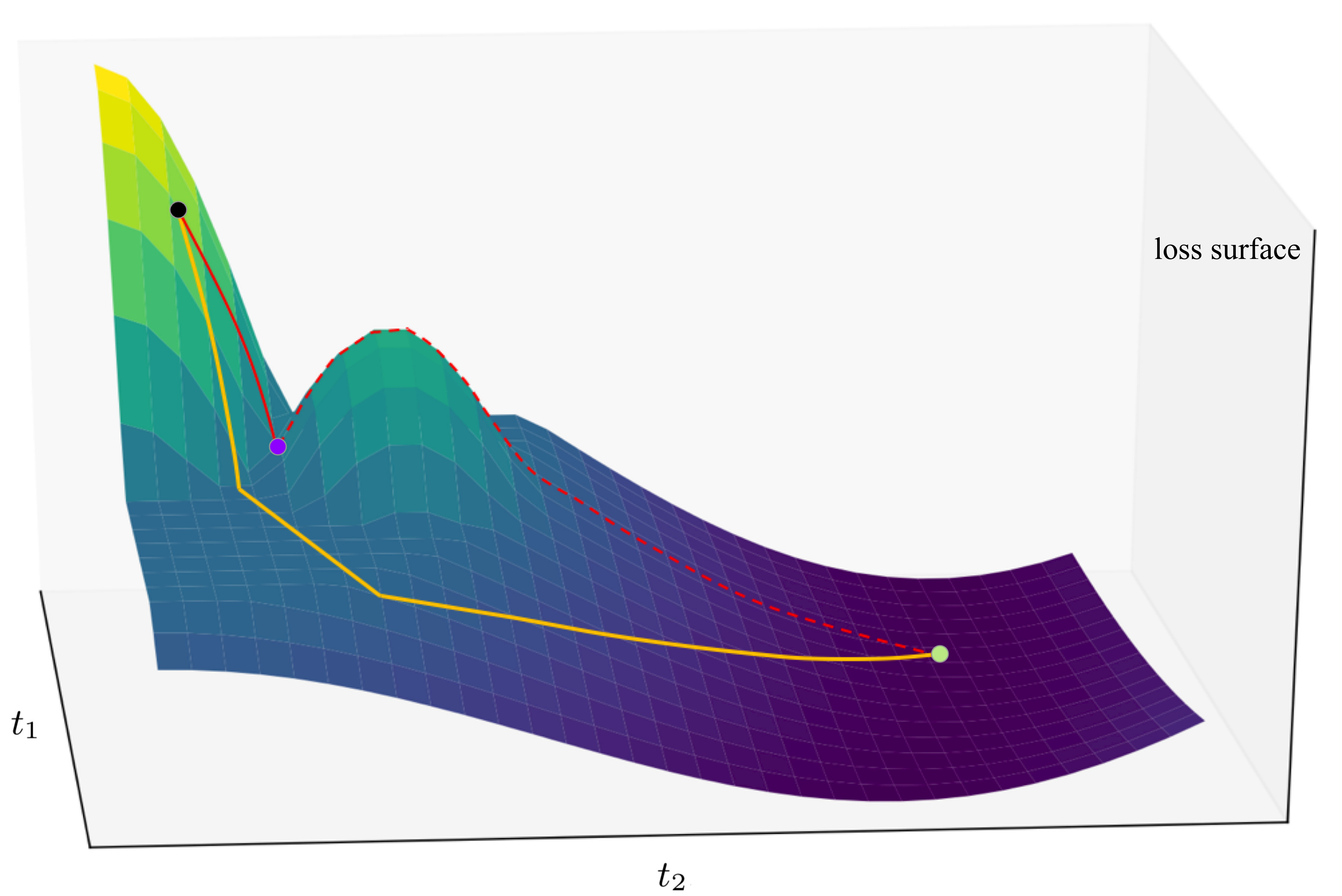}
			\scriptsize{(a). Loss surface}
		\end{minipage}
		\begin{minipage}[h]{0.45\linewidth}
			\centering
			\includegraphics[width=\linewidth]{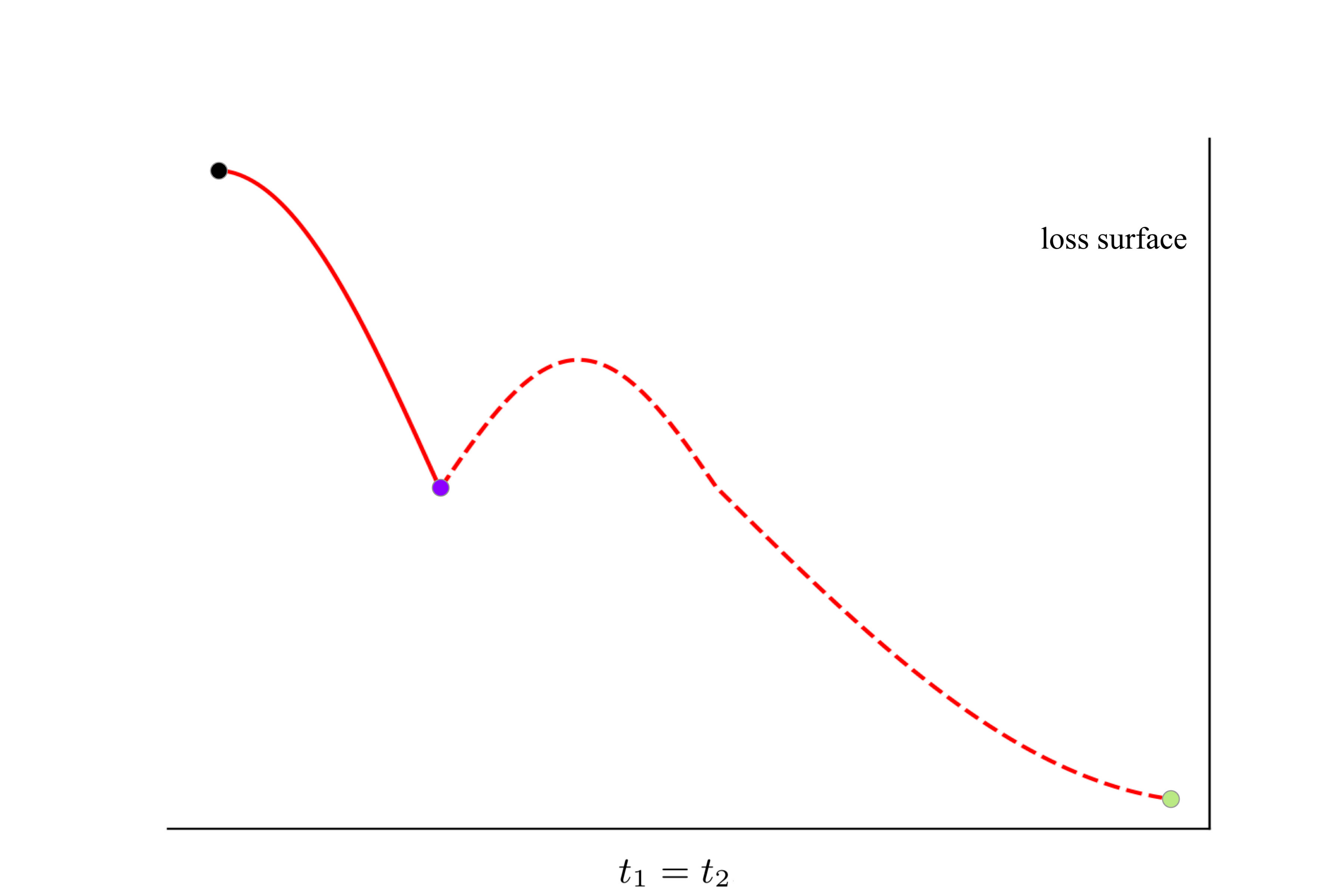}
			\scriptsize{(b). Constrained loss surface}
		\end{minipage}
		\caption{On the potential detrimental effect of using the atmosphere scattering model for image dehazing. For illustration purposes, we focus on two color channels of a single pixel and denote the respective transmission maps by $t_1$ and $t_2$. Fig.~\ref{fig:losssurface}(a) plots the loss surface as a function of $t_1$ and $t_2$. It can be seen that the global minimum is attained a point (see the green dot) satisfying $t_1=t_2$, which agrees with the atmosphere scattering model. With the black dot as the starting point, one can readily find this global minimum using gradient descent (see the yellow path). However, a restricted search based on the  atmosphere scattering model along the $t_1=t_2$ direction (see the red path) will get stuck at a point indicated by the purple dot (see Fig.~\ref{fig:losssurface}(b)). Note that this point is a local minimum in the constrained space but not in the original space, and it becomes an obstruction simply due to the adoption of the atmosphere scattering model.}
		\label{fig:losssurface}
	\end{figure}

	2. Trainable pre-processing module: 	
	The pre-processing module effectively converts the single image dehazing problem to a multi-image dehazing  problem by generating several variants of the given hazy image, each highlighting a different aspect of this image and making the relevant feature information more evidently exposed. 	In contrast to those hand-selected pre-processing methods adopted in the existing works ({\textit{e}.\textit{g}.,   \cite{iebmgated01}), the proposed pre-processing module is made fully trainable, which is in line with the general preference of  data-driven methods over prior-based methods as shown by recent developments in image dehazing.	
	Note that hand-selected processing methods typically aim to enhance certain concrete features that are visually recognizable. The exclusion of abstract features is not justifiable. Indeed, there might exist abstract transform domains that better suit the follow-up operations than the image domain.
	A trainable pre-processing module has the freedom to identify transform domains over which more diversity gain can be harnessed.

	3. Attention-based multi-scale estimation: Inspired by \cite{iebmgridnet01}, we implement multi-scale estimation on a grid network. The grid network  has clear advantages over the encoder-decoder network and the conventional multi-scale network extensively used in image restoration~\cite{iebmkpn01,iebmrdn01,Tao_2018_CVPR,iebmgated01}. In particular, the information flow in the encoder-decoder network or the conventional multi-scale network often suffers from the bottleneck effect due to the hierarchical architecture whereas the grid network circumvents this issue via dense connections across different scales using up-sampling/down-sampling blocks. We further endow the network with a channel-wise attention mechanism, which allows for  more flexible information exchange and aggregation. The attention mechanism also enables the network to better harness the diversity created by the pre-processing module.

\begin{figure*}[t]
	\centering
	\includegraphics[width=0.98\linewidth]{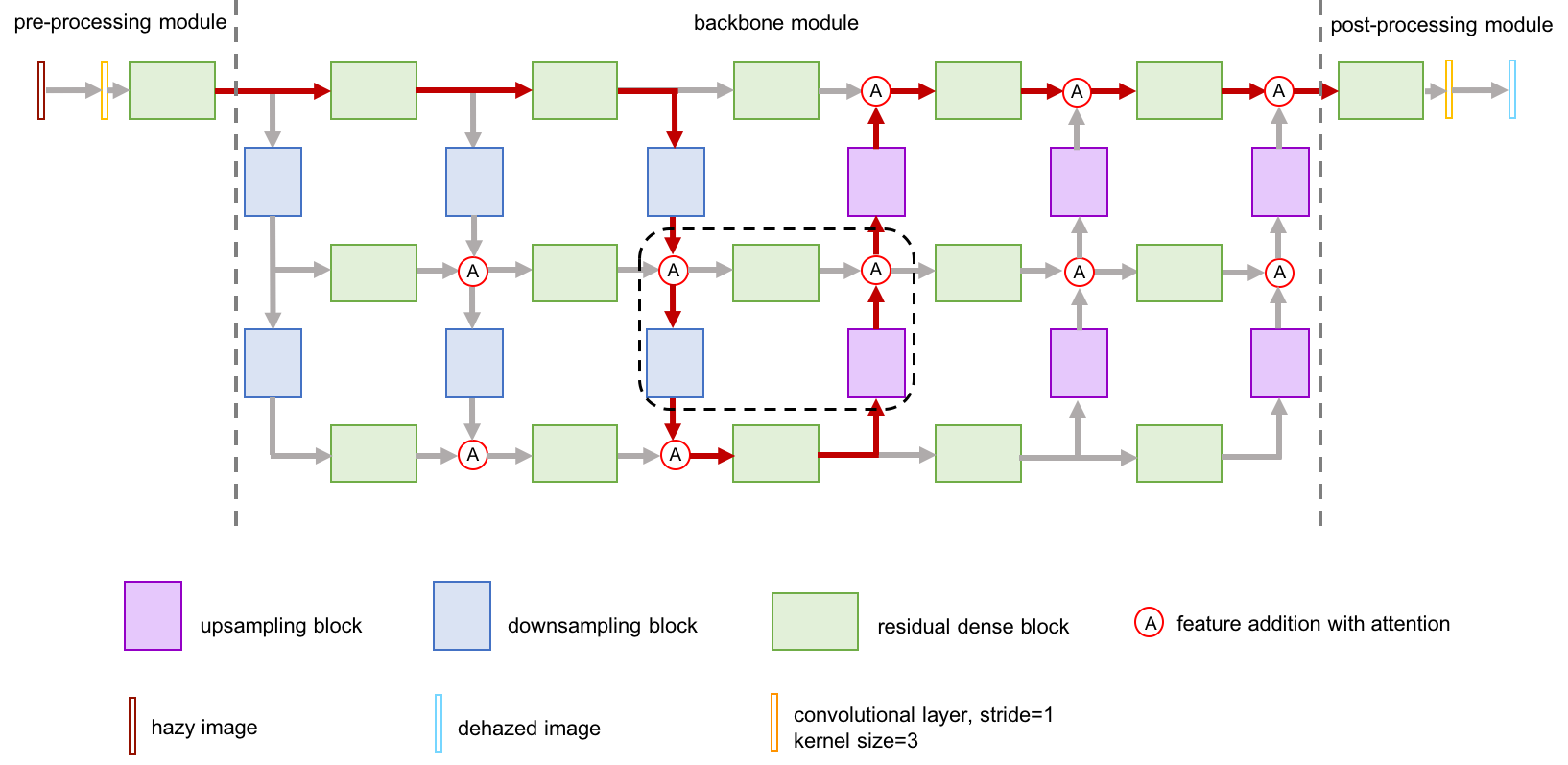}
	\caption{The architecture of GridDehazeNet.}
	\label{fig:GDN_main}
\end{figure*}

\begin{figure}[t]
	\centering
	\includegraphics[width=\linewidth]{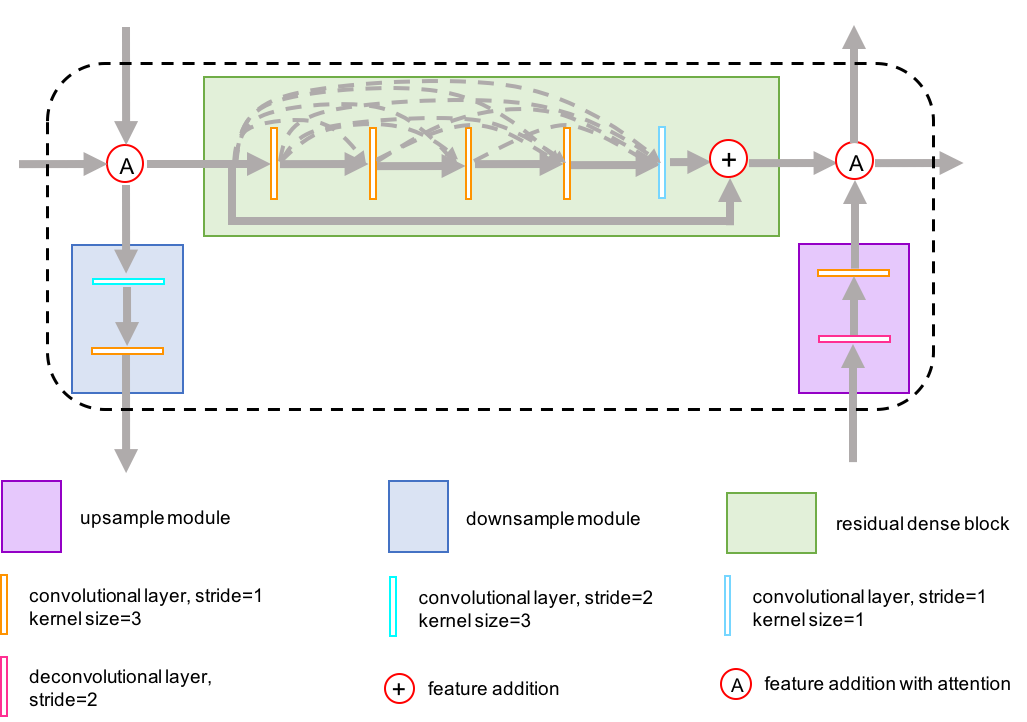}
	\caption{Illustration of the dash block in Fig.~\ref{fig:GDN_main}}
	\label{fig:GDN_block}
\end{figure}


\subsection{Network Architecture} \label{3.2}

The GridDehazeNet consists of three modules, namely, the pre-processing module, the backbone module and the post-processing module. 
Fig.~\ref{fig:GDN_main} shows the overall architecture of the proposed network. 

The pre-processing module consists of a convolutional layer  (w/o activation function) and a residual dense block (RDB)~\cite{iebmrdn01}. It generates 16 feature maps, which will be referred to as the learned inputs, from the given hazy image.



 The backbone module is an enhanced version of GridNet~\cite{iebmgridnet01} originally proposed for semantic segmentation. It performs attention-based multi-scale estimation based on the learned inputs generated by the pre-processing module. In this paper, we choose a grid network with three rows and six columns. Each row corresponds to a different scale and consists of five RDB blocks that keep the number of feature maps unchanged.  Each column can be regarded as a bridge that connects different scales via upsampling/downsampling blocks. In  each upsampling (downsampling) block, the size of feature maps is decreased (increased) by a factor of 2 while the number of feature maps is increased (decreased) by the same factor. Here upsampling/downsampling is realized using a convolutional layer instead of traditional methods such as bilinear or bicubic interpolation.  
 Fig.~\ref{fig:GDN_block} provides a detailed illustration of the RDB block, the upsampling block and the downsampling block. Each RDB block consists of five convolutional layers: the first four layers are used to increase the number of feature maps while the last layer fuses these feature maps and its output is then combined with the input of this RDB block via channel-wise addition. Following~\cite{iebmrdn01}, the growth rate in RDB is set to 16.  The upsampling block and the downsampling block are structurally the same except that different convolutional layers are used to adjust the size of feature maps. In the proposed GridDehazeNet, except for the first convolutional layer in the pre-processing module and the $1\times1$ convolutional layer in each RDB block, all convolutional layers employ ReLU as the activation function.  To strike a balance between the output size and the computational complexity, we set the number of feature maps at three different scales to 16, 32 and 64, respectively.
 
 


 The dehazed image constructed directly from the output of the backbone module tends to contain artifacts. As such, we introduce a post-processing module to improve the quality of the dehazed image. The structure of the post-processing module is symmetrical to that of the pre-processing module.

.


\subsection{Feature Fusion with Channel-Wise Attention} \label{3.3}


In view of the fact that feature maps from different scales may not be of the same importance, 
we propose a channel-wise attention mechanism, inspired by~\cite{iebmattention02}, to generate trainable weights for feature fusion.
Let $F_r^i$ and $F_c^i$ denote the $i$th feature channel from the row stream and the column stream, respectively, and let $a_r^i$ and $a_c^i$ denote their associated attention weights. The channel-wise attention mechanism can be expressed as 
\begin{equation}
\tilde{F}^i = a_r^iF_r^i + a_c^iF_c^i,
\end{equation}
where $\tilde{F}^i$ stands for the fused feature in the $i$th channel. The attention mechanism enables the GridDehazeNet to flexibly adjust the contributions from different scales in feature fusion. Our experimental results indicate that the performance of the proposed network can be greatly improved with the introduction of just a small number of trainable attention weights.

It is worth noting that one can prune (or deactivate) a portion of the proposed GridDehazeNet by choosing suitable attention weights and recover some existing network as a special case. For example, the red path in Fig.~\ref{fig:GDN_main}  illustrates an encoder-decoder network that can be obtained by pruning the GridDehazeNet. As another example, removing the exchange branches ({\textit{i}.\textit{e}.,  the middle four columns in the backbone module) from the GridDehazeNet leads to a structure resembling the conventional multi-scale network.



\subsection{Loss Function} \label{3.4}

To train the proposed network, the smooth $L_1$ loss and the perceptual loss~\cite{johnson2016perceptual} are employed. The smooth $L_1$ loss  provides a quantitative measure of the difference between the dehazed image and the ground truth, which is less sensitive to outliers than the MSE loss due to the fact that the $L_1$ norm can prevent potential gradient explosions~\cite{Girshick_2015_ICCV}.


\noindent\textbf{Smooth $L_1$ Loss}: Let $\hat{J}_i(x)$ denote the intensity of the $i$th color channel of pixel $x$ in the dehazed image, and $N$ denote the total number of pixels. The smooth $L_1$ Loss  can be expressed as
\begin{equation}
L_{S} = \frac{1}{N}\sum_{x=1}^N\sum_{i=1}^3F_{S}(\hat{J}_i(x)-J_i(x)),
\end{equation}
where 
\begin{eqnarray}F_{S}(e)=
\begin{cases}
0.5e^2, &\mbox{if }|e|<1,\cr |e|-0.5, &\mbox{otherwise}. \cr\end{cases}
\end{eqnarray}

\noindent\textbf{Perceptual Loss}: Different from the per-pixel loss, the perceptual loss leverages multi-scale features extracted from a  pre-trained deep neural network to quantify the visual difference between the estimated image and the ground truth. In this paper, we use the VGG16~\cite{simonyan2014very}  pre-trained on ImageNet~\cite{russakovsky2015imagenet} as the loss network  and extract the features from the last layer of each of the first three stages ({\textit{i}.\textit{e}.,  Conv1-2, Conv2-2 and Conv3-3). The perceptual loss is defined as 
\begin{equation}
L_{P} = \sum_{j=1}^3\frac{1}{C_jH_jW_j}||\phi_j(\hat{J})-\phi_j(J)||_2^2,
\end{equation}
where $\phi_j(\hat{J})$ ($\phi_j(J)$), $j=1,2,3$, denote the aforementioned three VGG16 feature maps associated with the dehazed image  $\hat{J}$ (the ground truth $J$), and $C_j$, $H_j$ and $W_j$ specify the dimension of $\phi_j(\hat{J})$ ($\phi_j(J)$), $j=1,2,3$.



\noindent\textbf{Total Loss}: The total loss is defined by combining the smooth $L_1$ loss and the perceptual loss as follows:
\begin{equation}
L = L_S + \lambda L_P,
\end{equation}
where $\lambda$ is a parameter used to adjust the relative weights on the two loss components.
In this paper, $\lambda$ is set to 0.04.

\section{Experimental Results}


We conduct extensive experiments to demonstrate that the proposed GridDehazeNet performs favorably against the state-of-the-arts in terms of quantitative dehazing results and  qualitative visual effects on synthetic and real-world datasets. The experimental results also provide useful insights into the constituent modules of GridDehazeNet and solid justifications for the overall design. More examples can be found in the supplementary material and the source code will be made publicly available.



 
\subsection{Training and Testing Dataset}
In general it is impractical to collect a large number of real-world hazy images and their haze-free counterparts. Therefore, data-driven dehazing methods often need to rely on
synthetic hazy images, which can be generated from clear images based on the atmosphere scattering model via proper choice of the scattering coefficient $\beta$ and  the atmospheric light intensity $A$. In this paper, we adopt a large-scale synthetic dataset, named RESIDE~\cite{li2019benchmarking}, to train and test the proposed GridDehazeNet. RESIDE contains synthetic hazy images in both indoor and outdoor scenarios. The Indoor Training Set (ITS) of RESIDE contains
a total of 13990   hazy indoor images, generated from 1399 clear images with $\beta \in [0.6, 1.8]$ and $A \in [0.7, 1.0]$; the depth maps $d(x)$ are obtained 
from the NYU Depth V2~\cite{silberman2012indoor} and Middlebury Stereo datasets~\cite{scharstein2003high}. 
After data cleaning, the Outdoor Training Set (OTS) of RESIDE contains a total of 296695  hazy outdoor images,
generated from 8477 clear images with $\beta \in [0.04, 0.2]$ and $A \in [0.8, 1.0]$; the depth maps of outdoor images are estimated using the algorithm developed in~\cite{liu2016learning}.   
For testing, the Synthetic Objective Testing Set (SOTS) is adopted, which consists of 500 indoor hazy images and 500 outdoor ones. Moreover, for comparisons on real-world images, we use the dataset from \cite{fattal2014dehazing}.

\subsection{Implementation}\label{sec:implementation}
The proposed GridDehazeNet is end-to-end trainable without the need of pre-training for sub-modules. 
We train the network with RGB image patches of size $240 \times 240$. For accelerated training, the Adam optimizer~\cite{kingma2014adam} is used with a batch size of 24, where $\beta_1$ and $\beta_2$ take the default values of 0.9 and 0.999, respectively.  Following~\cite{nah2017deep, lim2017enhanced}, we do not use batch normalization.
The initial learning rate is set to 0.001. For ITS, we train the network for 100 epochs in total and reduce the learning rate by half every 20 epochs. As for OTS, the network is trained only for 10 epochs and the learning rate is reduced by half every 2 epochs. The training is carried out on a PC with two NVIDIA GeForce GTX 1080Ti, but only one GPU is used for testing. When the training ends, the loss functions for ITS and OTS drop to 0.0005 and 0.0004, respectively, which we consider as a good indication of convergence.



\begin{figure*}[t]
	\centering
	\begin{minipage}[h]{0.12\linewidth}
		\centering
		\includegraphics[width=\linewidth]{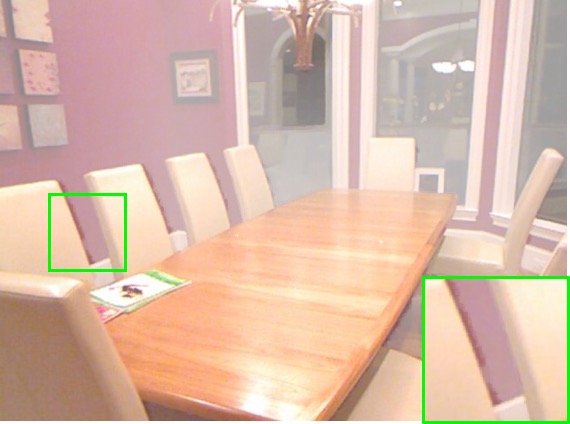}
	\end{minipage}
	\begin{minipage}[h]{0.12\linewidth}
		\centering
		\includegraphics[width=\linewidth]{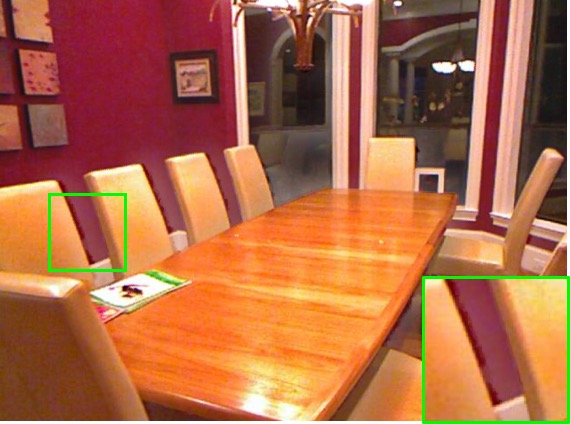}
	\end{minipage}
	\begin{minipage}[h]{0.12\linewidth}
		\centering
		\includegraphics[width=\linewidth]{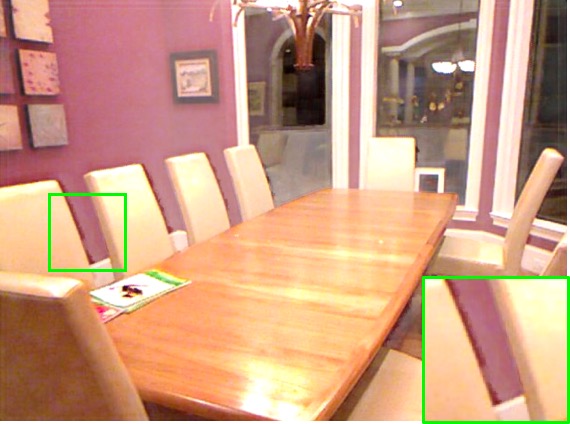}
	\end{minipage}
	\begin{minipage}[h]{0.12\linewidth}
		\centering
		\includegraphics[width=\linewidth]{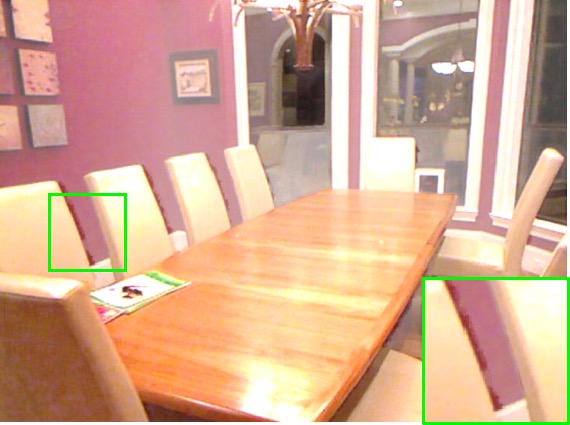}
	\end{minipage}
	\begin{minipage}[h]{0.12\linewidth}
		\centering
		\includegraphics[width=\linewidth]{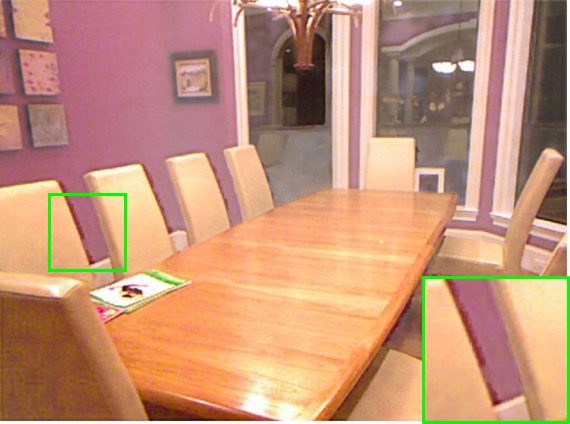}
	\end{minipage}
	\begin{minipage}[h]{0.12\linewidth}
		\centering
		\includegraphics[width=\linewidth]{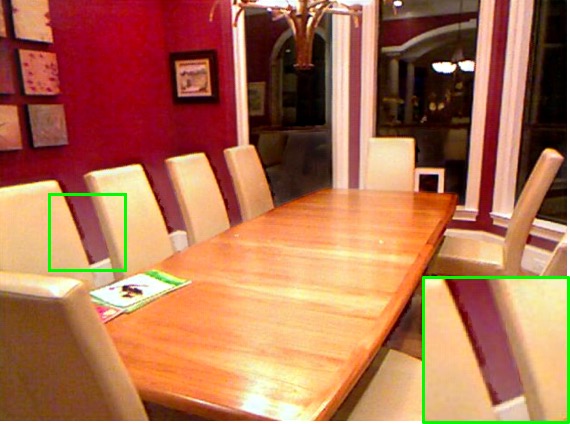}
	\end{minipage}
	\begin{minipage}[h]{0.12\linewidth}
		\centering
		\includegraphics[width=\linewidth]{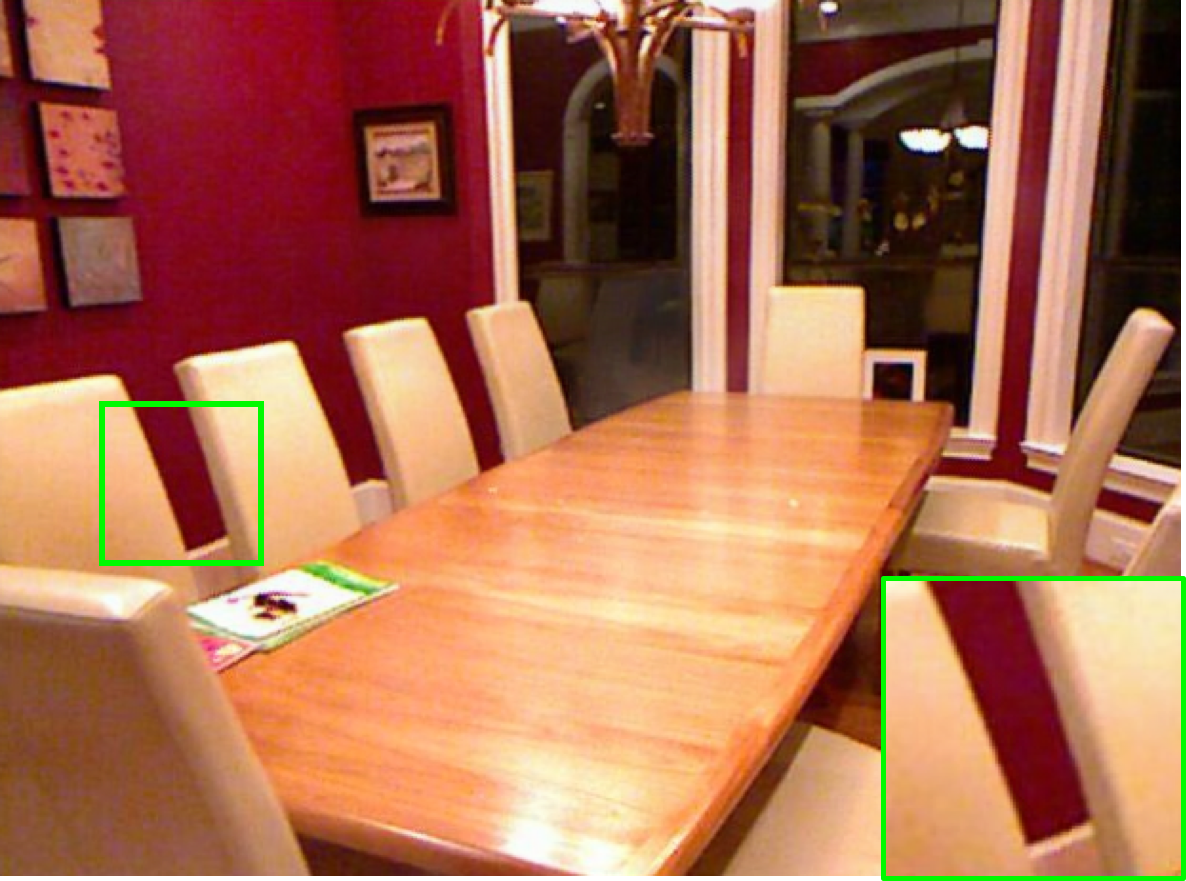}
	\end{minipage}
	\begin{minipage}[h]{0.12\linewidth}
		\centering
		\includegraphics[width=\linewidth]{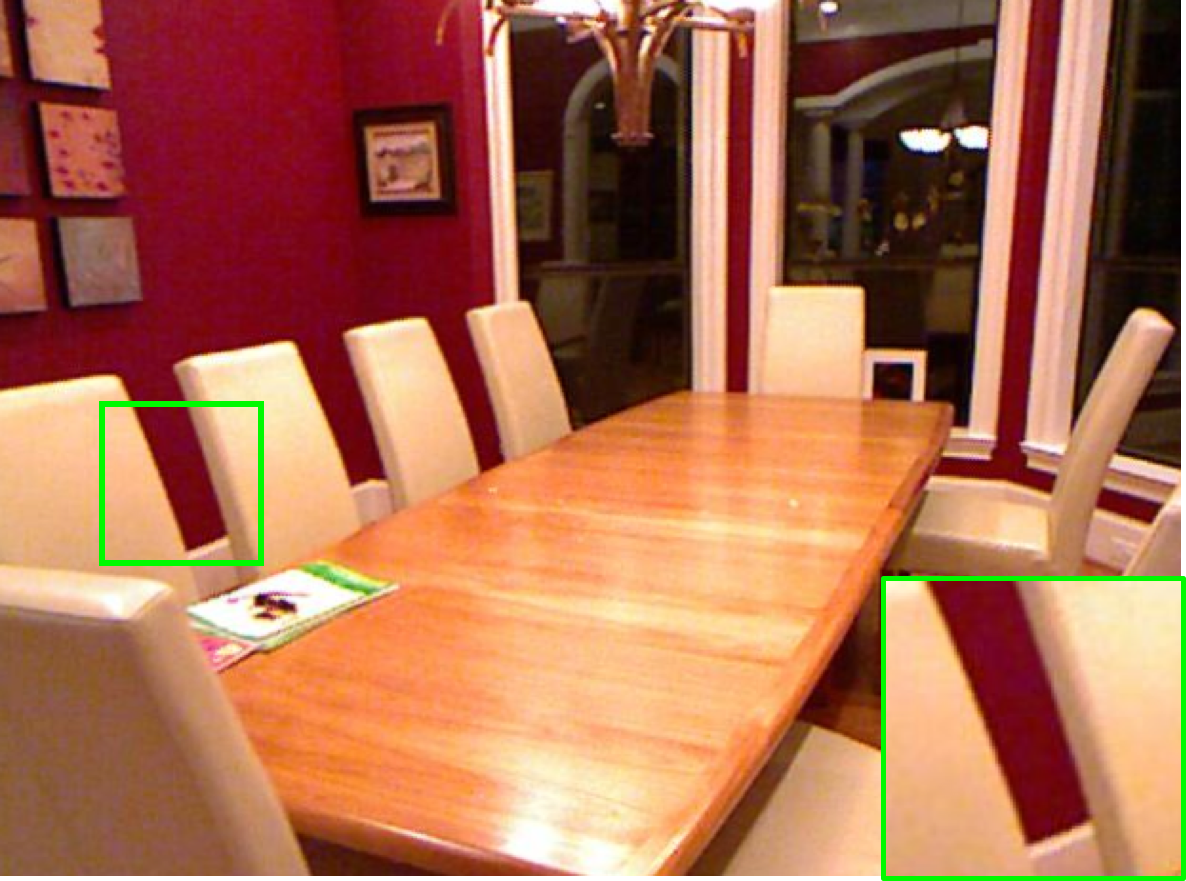}
	\end{minipage}
	\begin{minipage}[h]{0.12\linewidth}
		\centering
		\includegraphics[width=\linewidth]{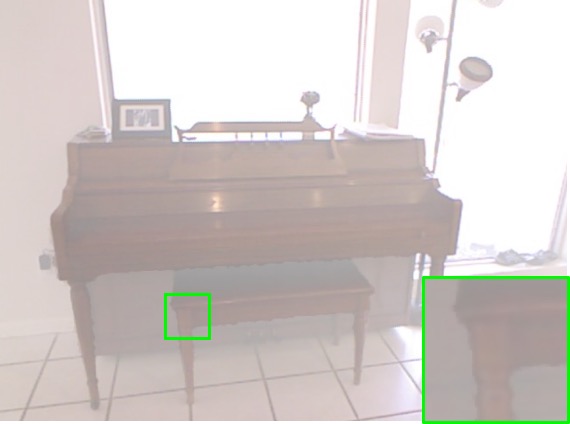}
	\end{minipage}
	\begin{minipage}[h]{0.12\linewidth}
		\centering
		\includegraphics[width=\linewidth]{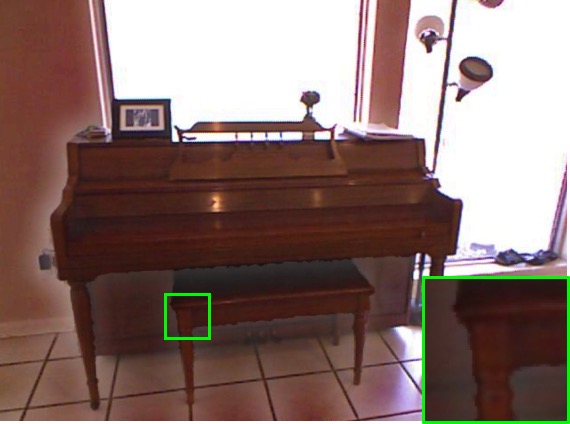}
	\end{minipage}
	\begin{minipage}[h]{0.12\linewidth}
		\centering
		\includegraphics[width=\linewidth]{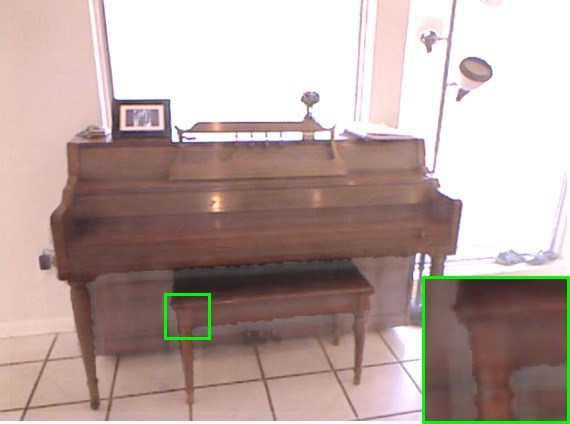}
	\end{minipage}
	\begin{minipage}[h]{0.12\linewidth}
		\centering
		\includegraphics[width=\linewidth]{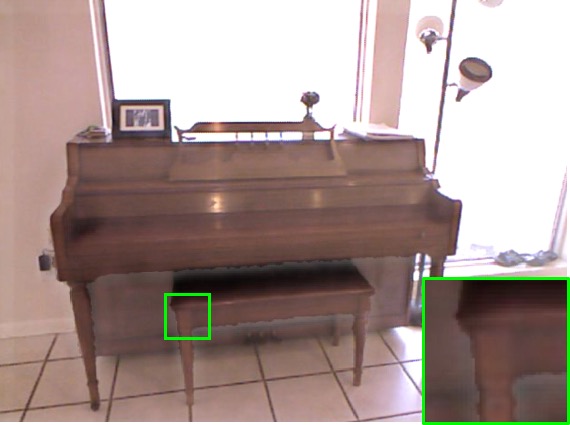}
	\end{minipage}
	\begin{minipage}[h]{0.12\linewidth}
		\centering
		\includegraphics[width=\linewidth]{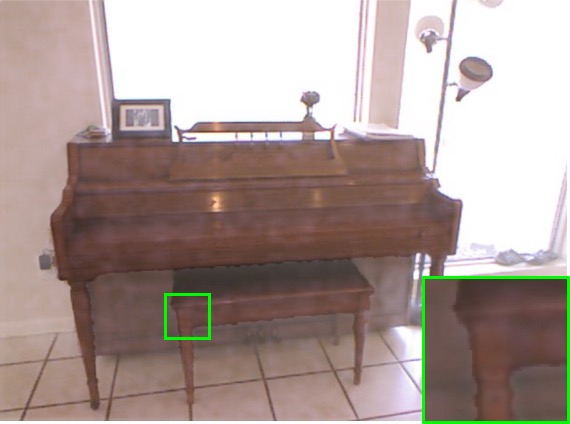}
	\end{minipage}
	\begin{minipage}[h]{0.12\linewidth}
		\centering
		\includegraphics[width=\linewidth]{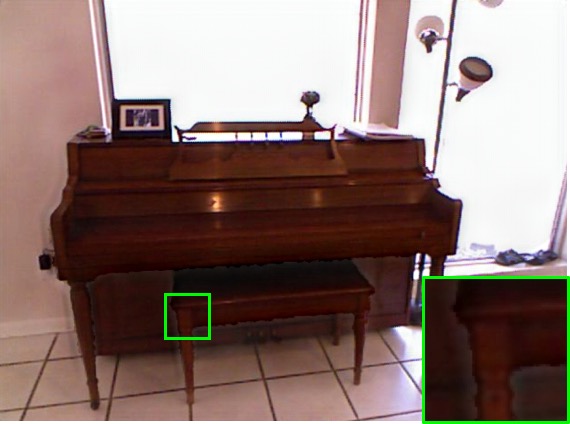}
	\end{minipage}
	\begin{minipage}[h]{0.12\linewidth}
		\centering
		\includegraphics[width=\linewidth]{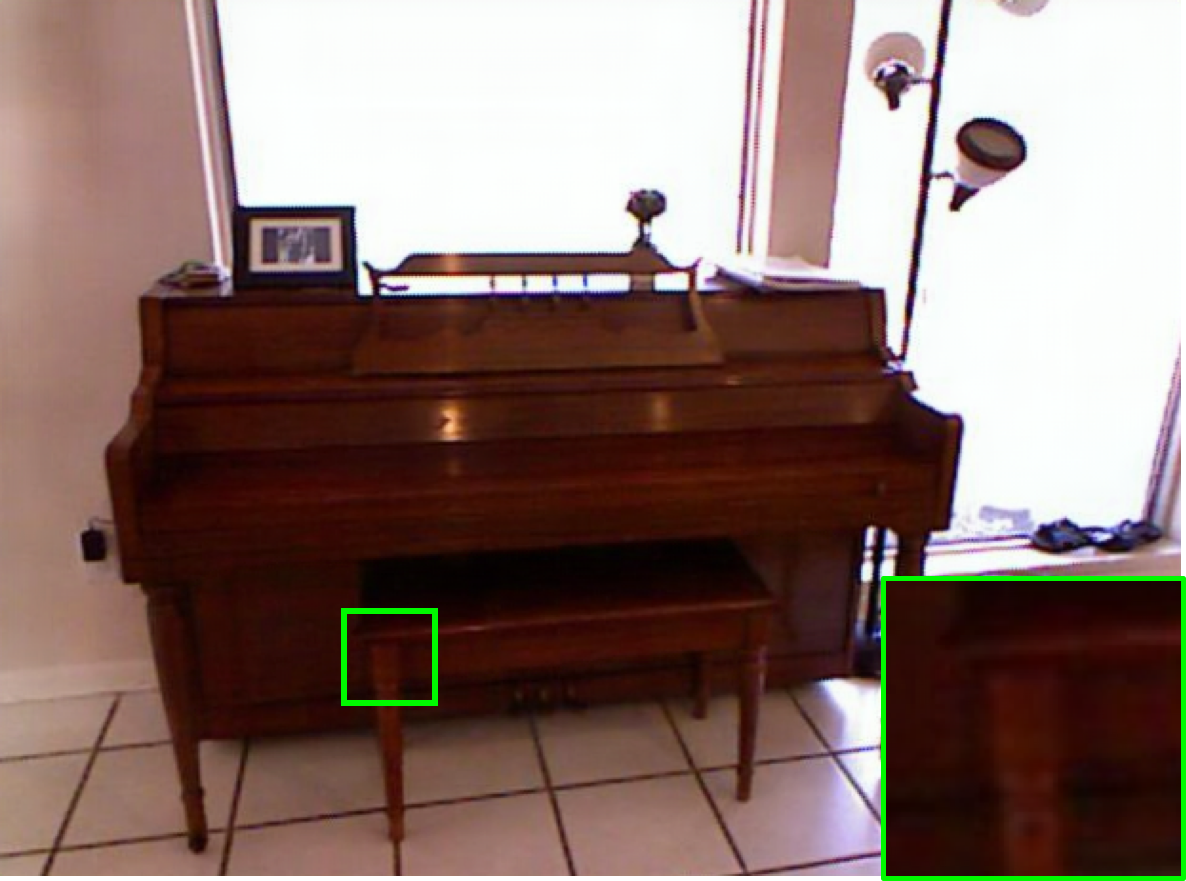}
	\end{minipage}
	\begin{minipage}[h]{0.12\linewidth}
		\centering
		\includegraphics[width=\linewidth]{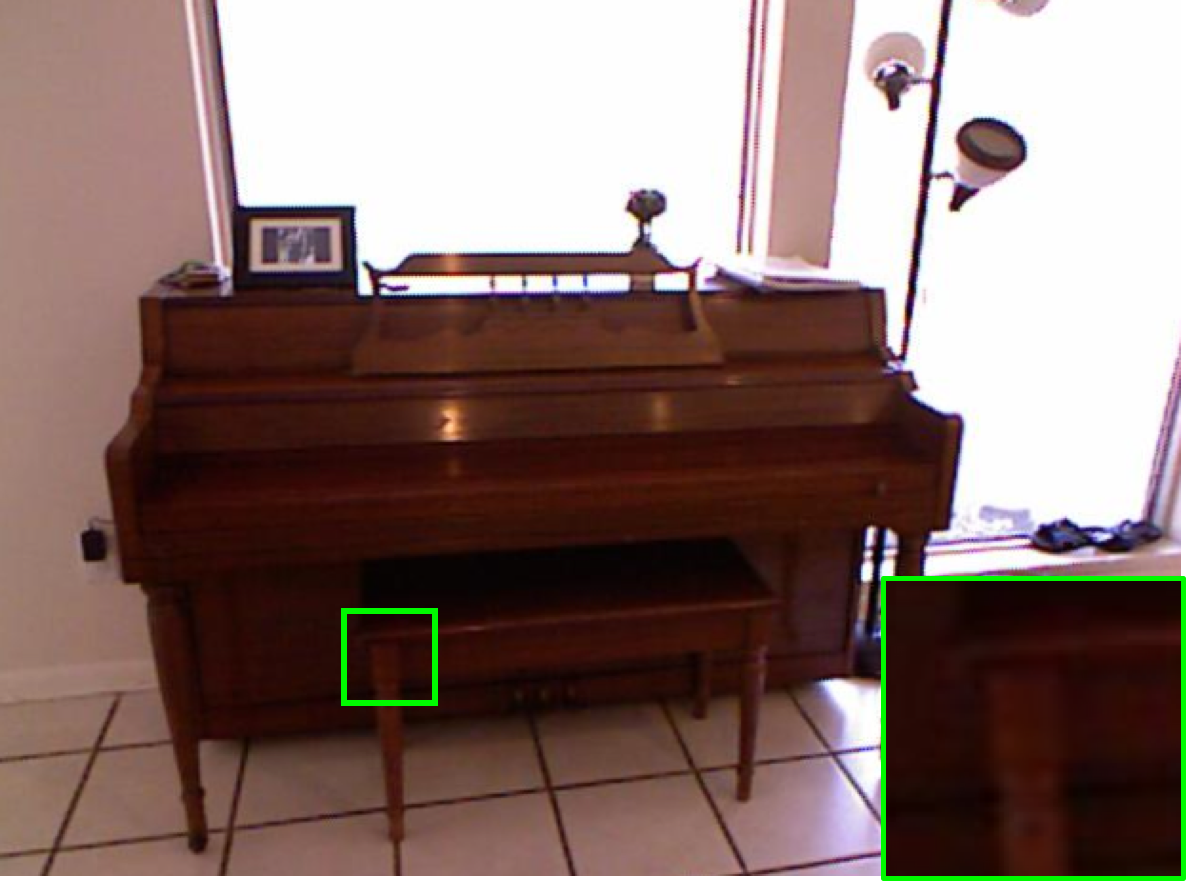}
	\end{minipage}
	
	\begin{minipage}[h]{0.12\linewidth}
		\centering
		\includegraphics[width=\linewidth]{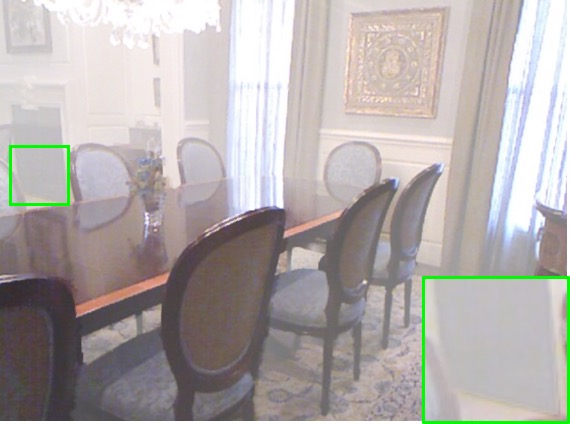}
	\end{minipage}
	\begin{minipage}[h]{0.12\linewidth}
		\centering
		\includegraphics[width=\linewidth]{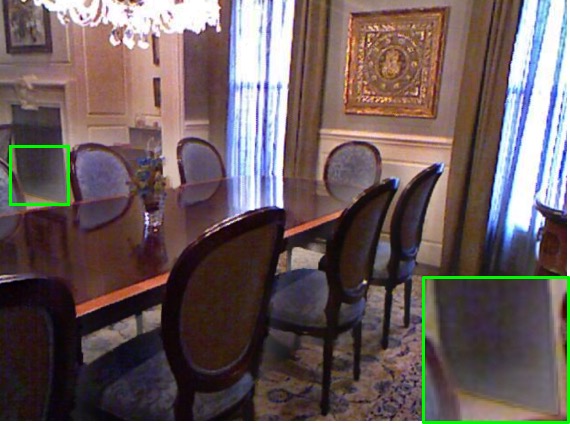}
	\end{minipage}
	\begin{minipage}[h]{0.12\linewidth}
		\centering
		\includegraphics[width=\linewidth]{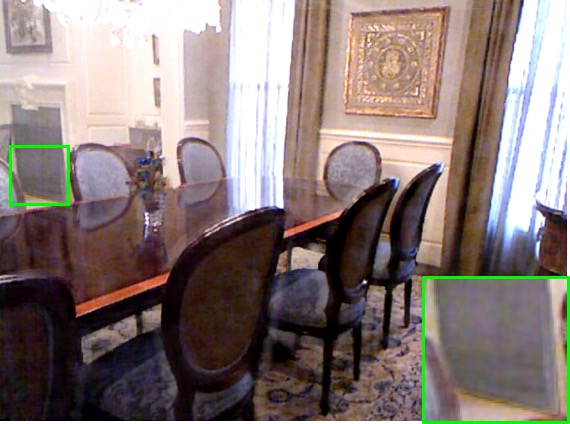}
	\end{minipage}
	\begin{minipage}[h]{0.12\linewidth}
		\centering
		\includegraphics[width=\linewidth]{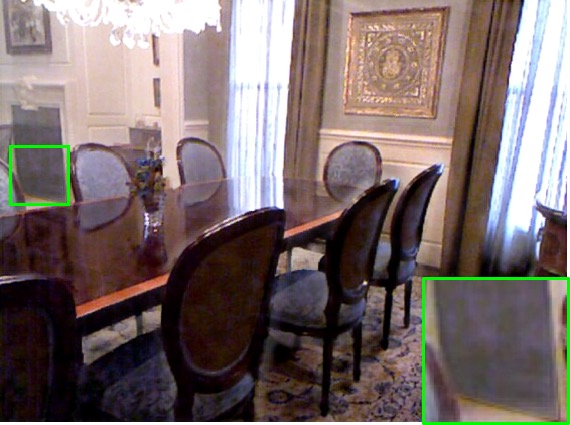}
	\end{minipage}
	\begin{minipage}[h]{0.12\linewidth}
		\centering
		\includegraphics[width=\linewidth]{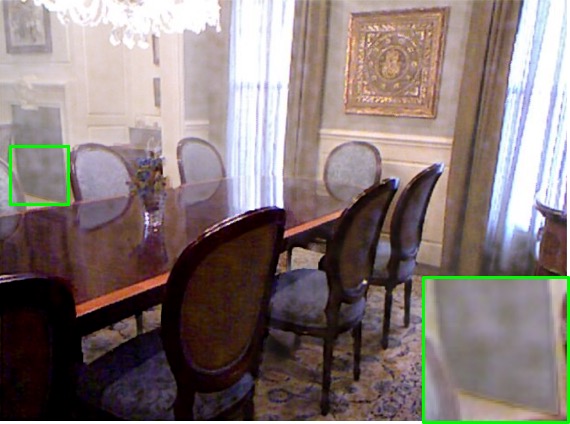}
	\end{minipage}
	\begin{minipage}[h]{0.12\linewidth}
		\centering
		\includegraphics[width=\linewidth]{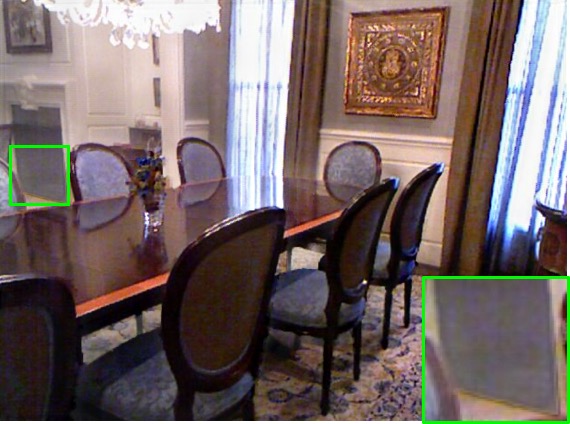}
	\end{minipage}
	\begin{minipage}[h]{0.12\linewidth}
		\centering
		\includegraphics[width=\linewidth]{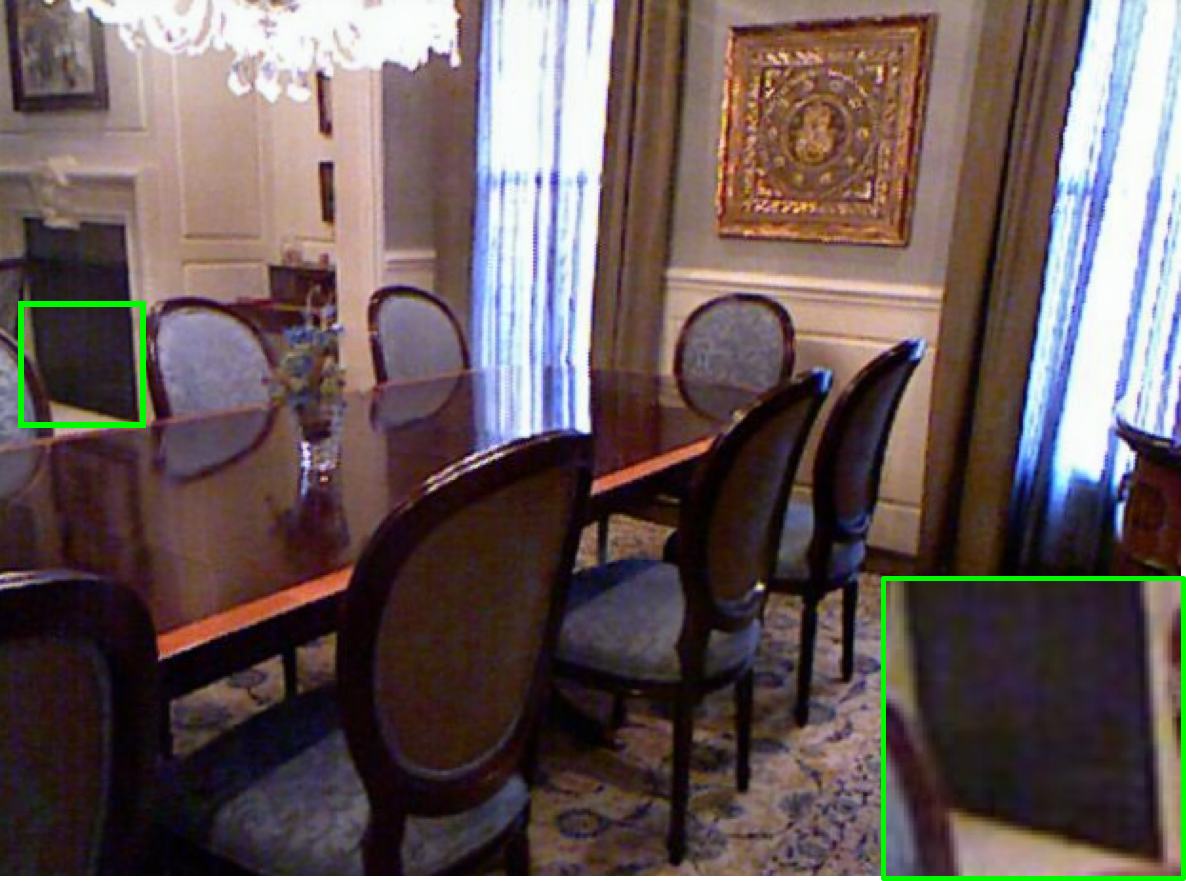}
	\end{minipage}
	\begin{minipage}[h]{0.12\linewidth}
		\centering
		\includegraphics[width=\linewidth]{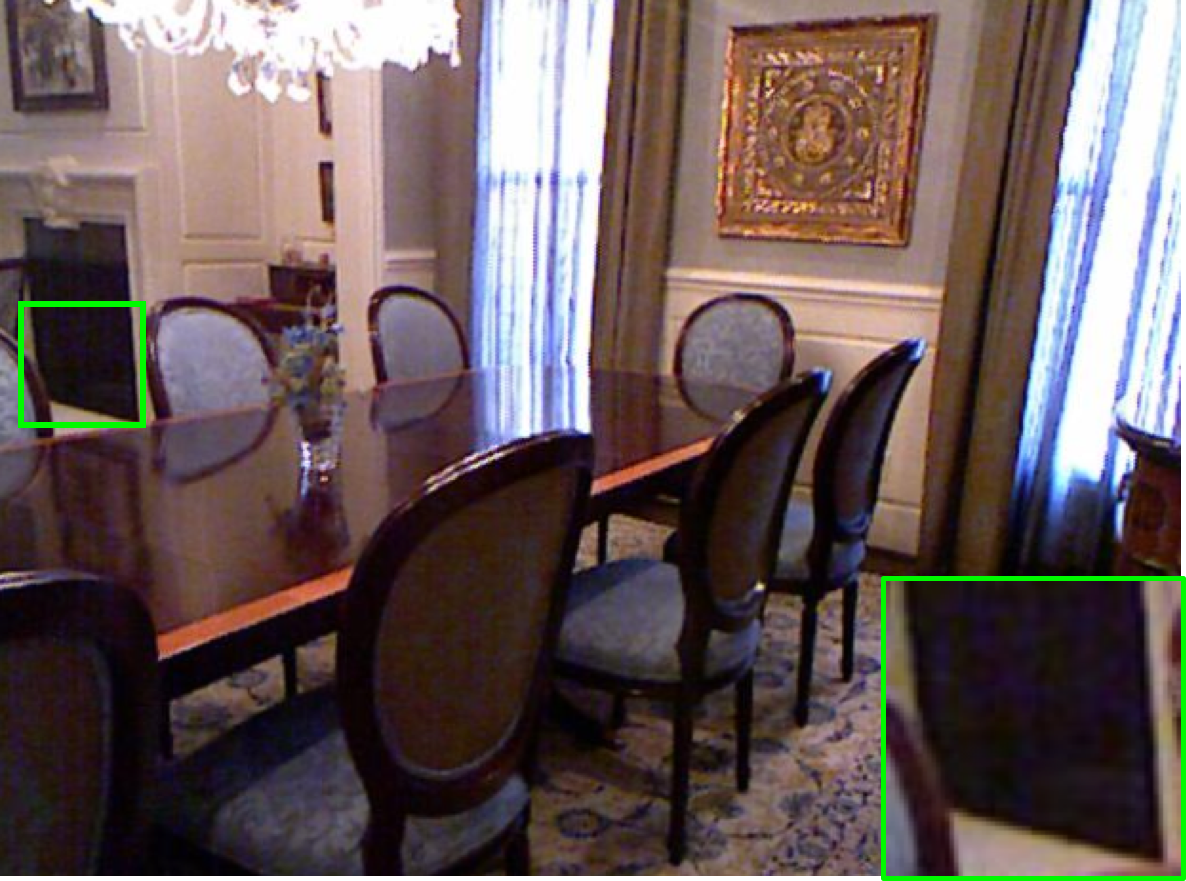}
	\end{minipage}
	
	\begin{minipage}[h]{0.12\linewidth}
		\centering
		\includegraphics[width=\linewidth]{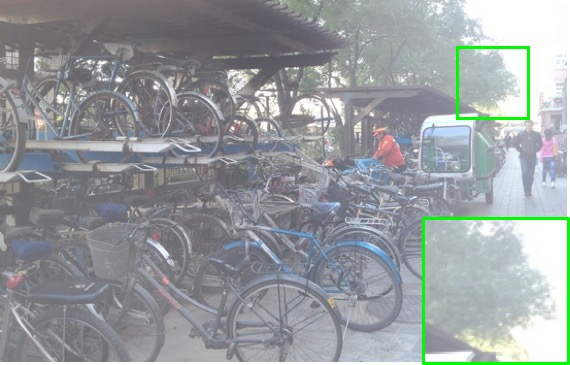}
	\end{minipage}
	\begin{minipage}[h]{0.12\linewidth}
		\centering
		\includegraphics[width=\linewidth]{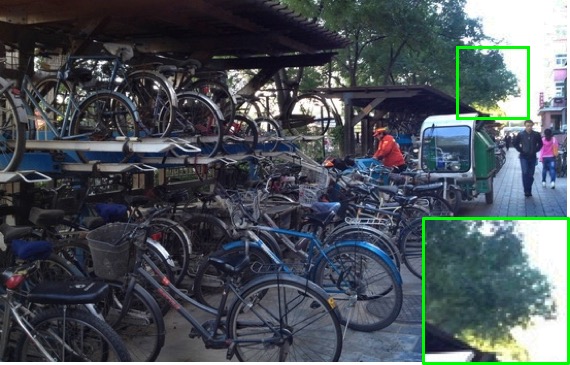}
	\end{minipage}
	\begin{minipage}[h]{0.12\linewidth}
		\centering
		\includegraphics[width=\linewidth]{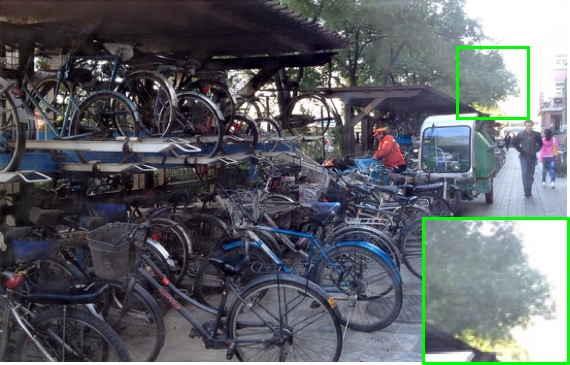}
	\end{minipage}
	\begin{minipage}[h]{0.12\linewidth}
		\centering
		\includegraphics[width=\linewidth]{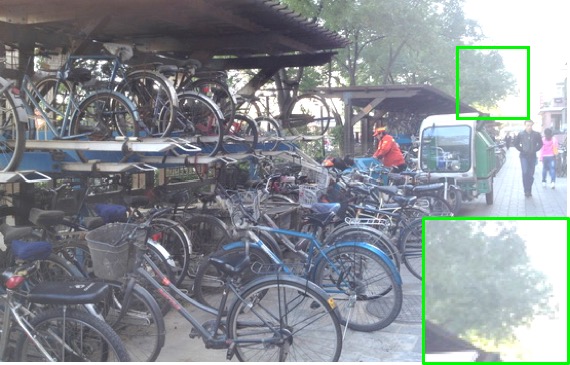}
	\end{minipage}
	\begin{minipage}[h]{0.12\linewidth}
		\centering
		\includegraphics[width=\linewidth]{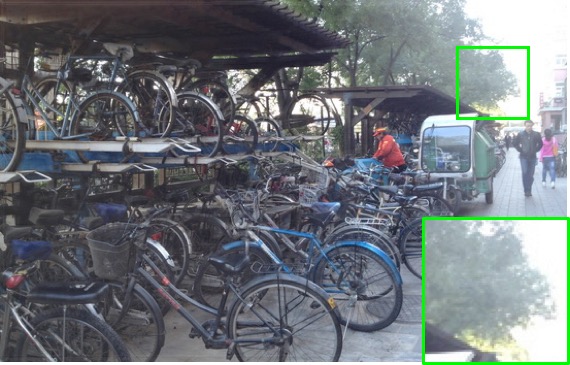}
	\end{minipage}
	\begin{minipage}[h]{0.12\linewidth}
		\centering
		\includegraphics[width=\linewidth]{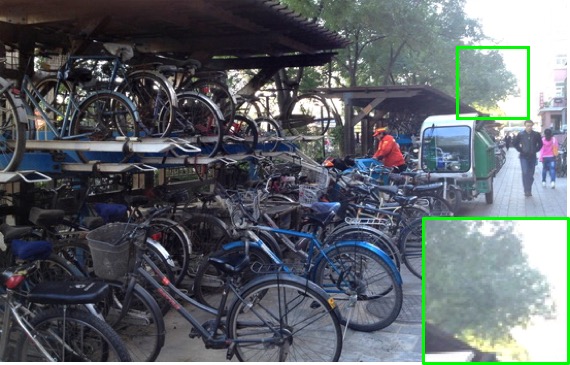}
	\end{minipage}
	\begin{minipage}[h]{0.12\linewidth}
		\centering
		\includegraphics[width=\linewidth]{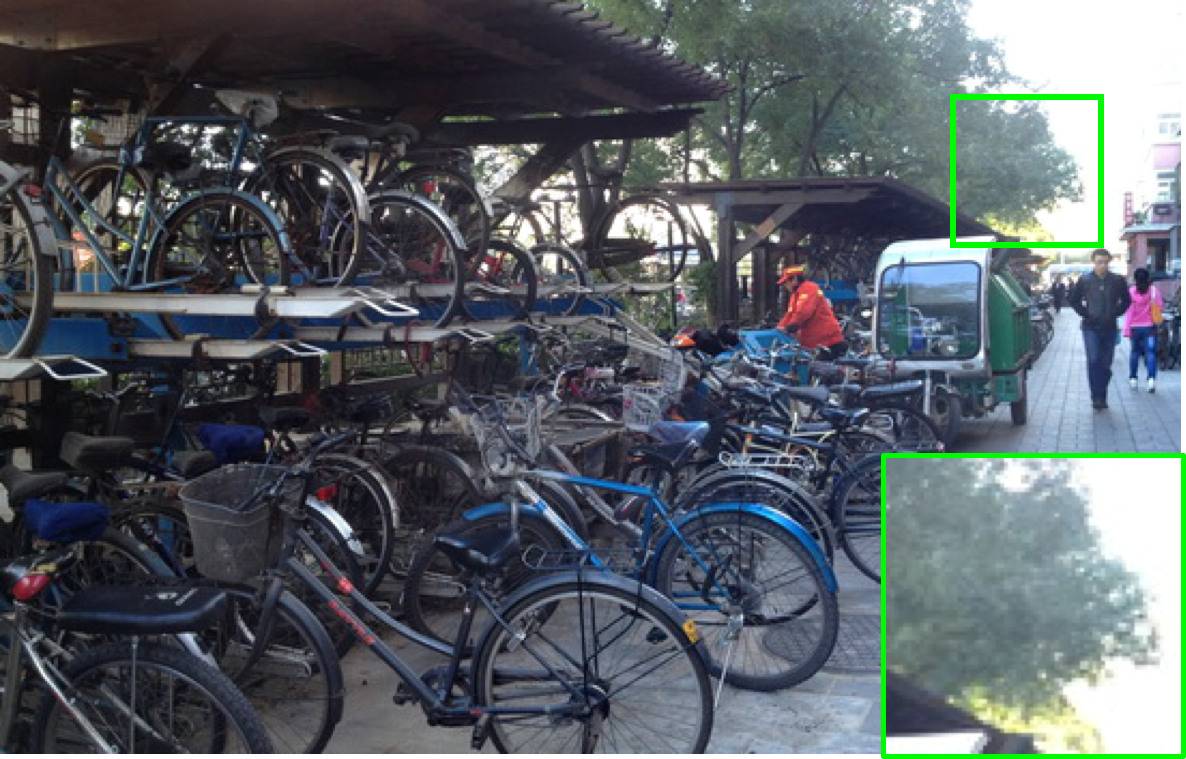}
	\end{minipage}
	\begin{minipage}[h]{0.12\linewidth}
		\centering
		\includegraphics[width=\linewidth]{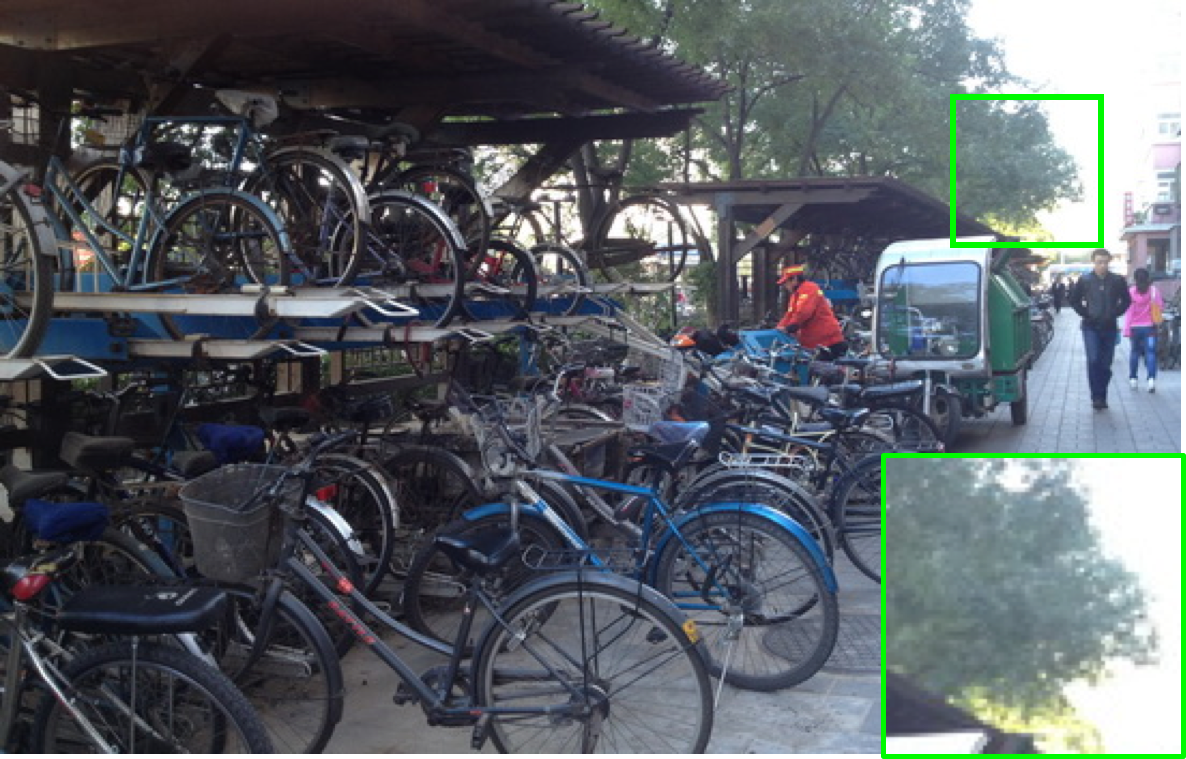}
	\end{minipage}
	
	\begin{minipage}[h]{0.12\linewidth}
		\centering
		\includegraphics[width=\linewidth]{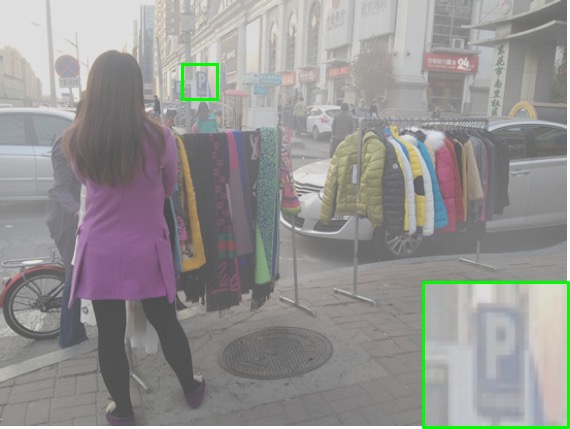}
	\end{minipage}
	\begin{minipage}[h]{0.12\linewidth}
		\centering
		\includegraphics[width=\linewidth]{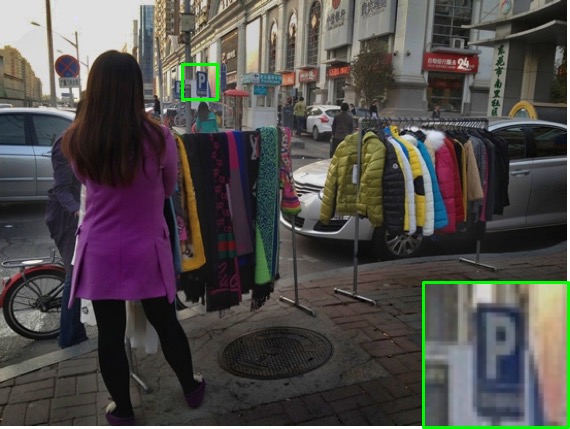}
	\end{minipage}
	\begin{minipage}[h]{0.12\linewidth}
		\centering
		\includegraphics[width=\linewidth]{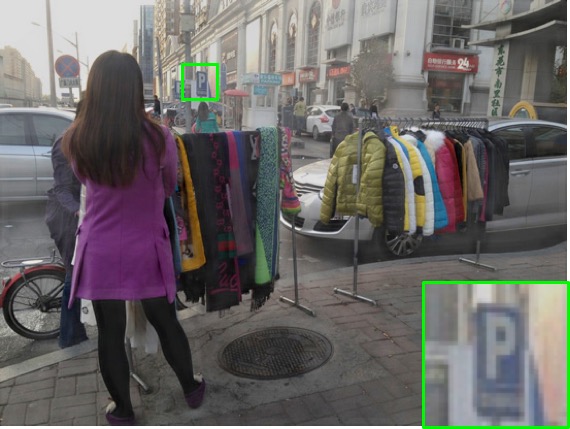}
	\end{minipage}
	\begin{minipage}[h]{0.12\linewidth}
		\centering
		\includegraphics[width=\linewidth]{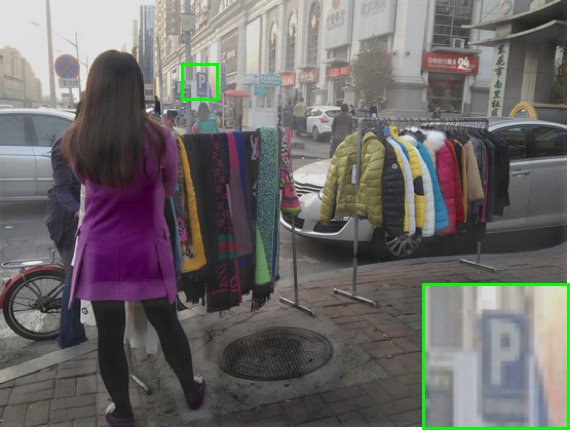}
	\end{minipage}
	\begin{minipage}[h]{0.12\linewidth}
		\centering
		\includegraphics[width=\linewidth]{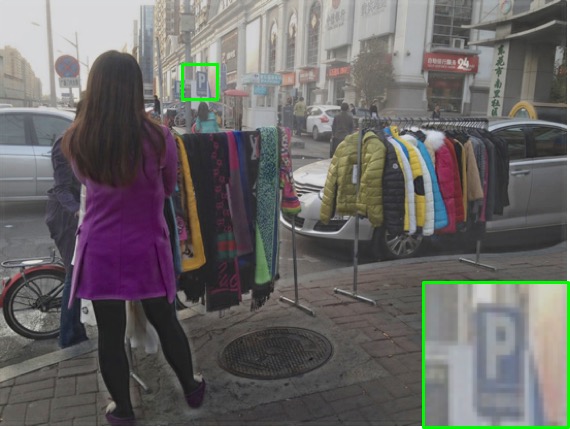}
	\end{minipage}
	\begin{minipage}[h]{0.12\linewidth}
		\centering
		\includegraphics[width=\linewidth]{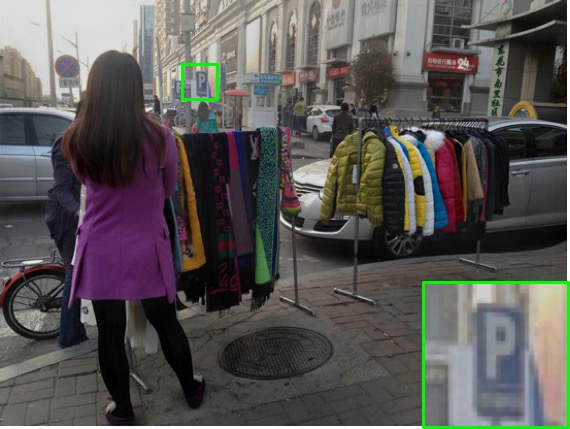}
	\end{minipage}
	\begin{minipage}[h]{0.12\linewidth}
		\centering
		\includegraphics[width=\linewidth]{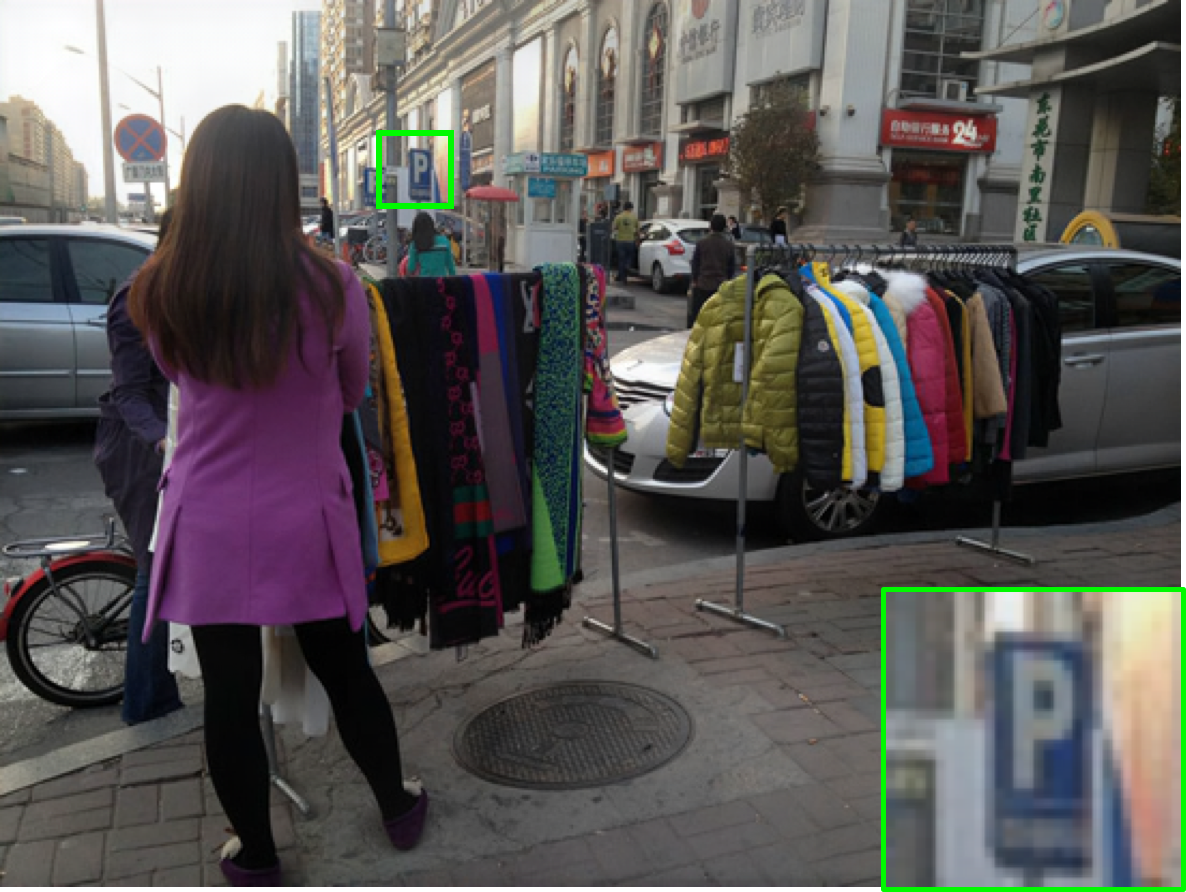}
	\end{minipage}
	\begin{minipage}[h]{0.12\linewidth}
		\centering
		\includegraphics[width=\linewidth]{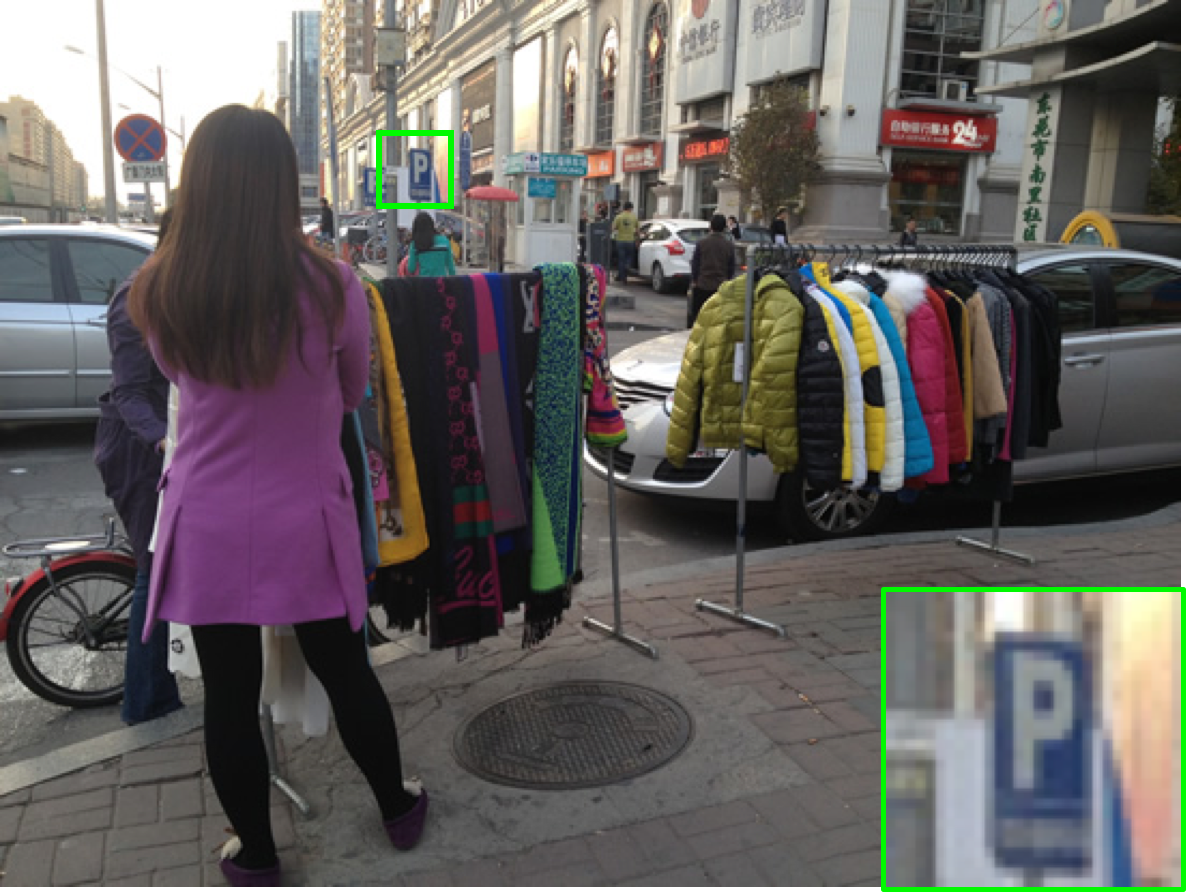}
	\end{minipage}
	\begin{minipage}[h]{0.12\linewidth}
		\centering
		\includegraphics[width=\linewidth]{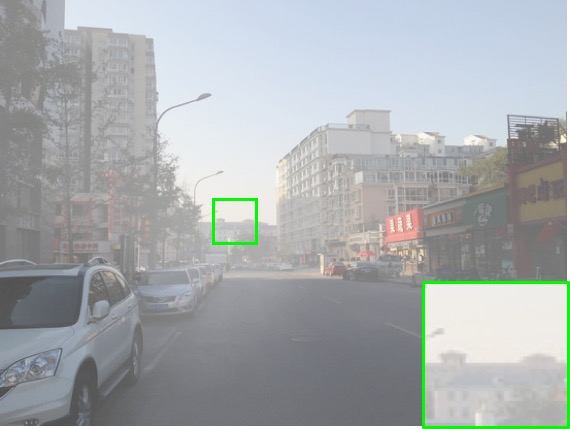}
		\scriptsize{(a) Hazy inputs}
	\end{minipage}
	\begin{minipage}[h]{0.12\linewidth}
		\centering
		\includegraphics[width=\linewidth]{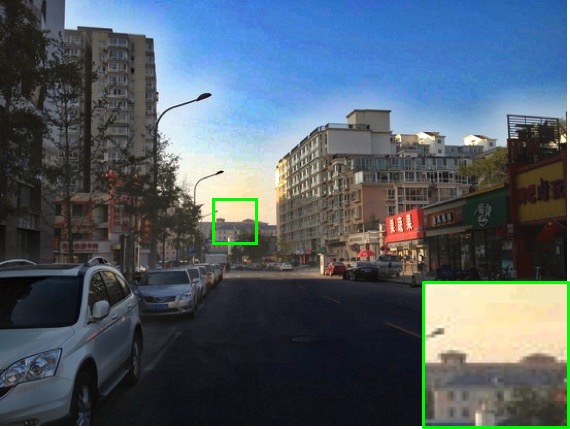}
		\scriptsize{(b) DCP}
	\end{minipage}
	\begin{minipage}[h]{0.12\linewidth}
		\centering
		\includegraphics[width=\linewidth]{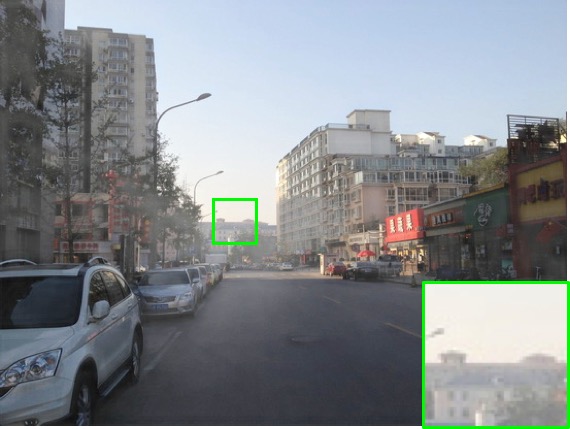}
		\scriptsize{(c) DehazeNet}
	\end{minipage}
	\begin{minipage}[h]{0.12\linewidth}
		\centering
		\includegraphics[width=\linewidth]{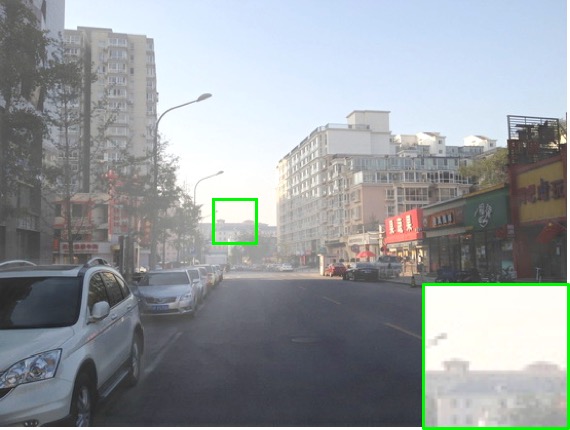}
		\scriptsize{(d) MSCNN}
	\end{minipage}
	\begin{minipage}[h]{0.12\linewidth}
		\centering
		\includegraphics[width=\linewidth]{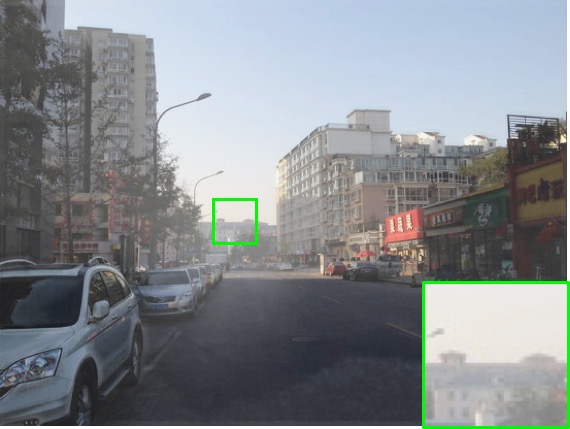}
		\scriptsize{(e) AOD-Net}
	\end{minipage}
	\begin{minipage}[h]{0.12\linewidth}
		\centering
		\includegraphics[width=\linewidth]{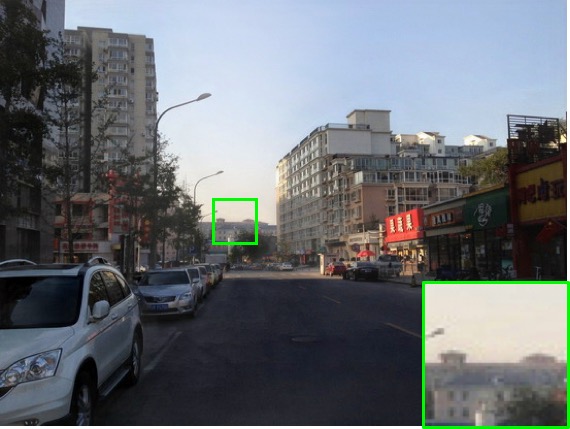}
		\scriptsize{(f) GFN}
	\end{minipage}
	\begin{minipage}[h]{0.12\linewidth}
		\centering
		\includegraphics[width=\linewidth]{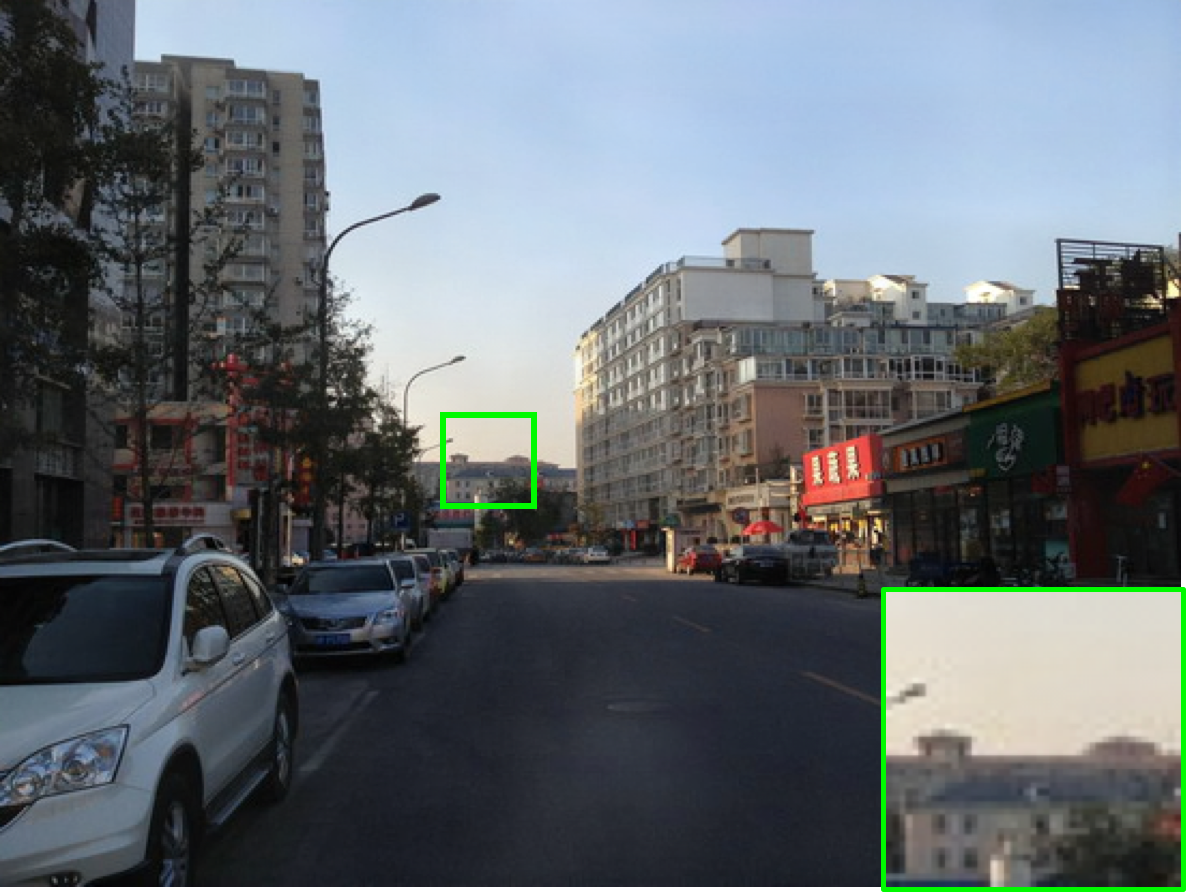}
		\scriptsize{(g) Ours}
	\end{minipage}
	\begin{minipage}[h]{0.12\linewidth}
		\centering
		\includegraphics[width=\linewidth]{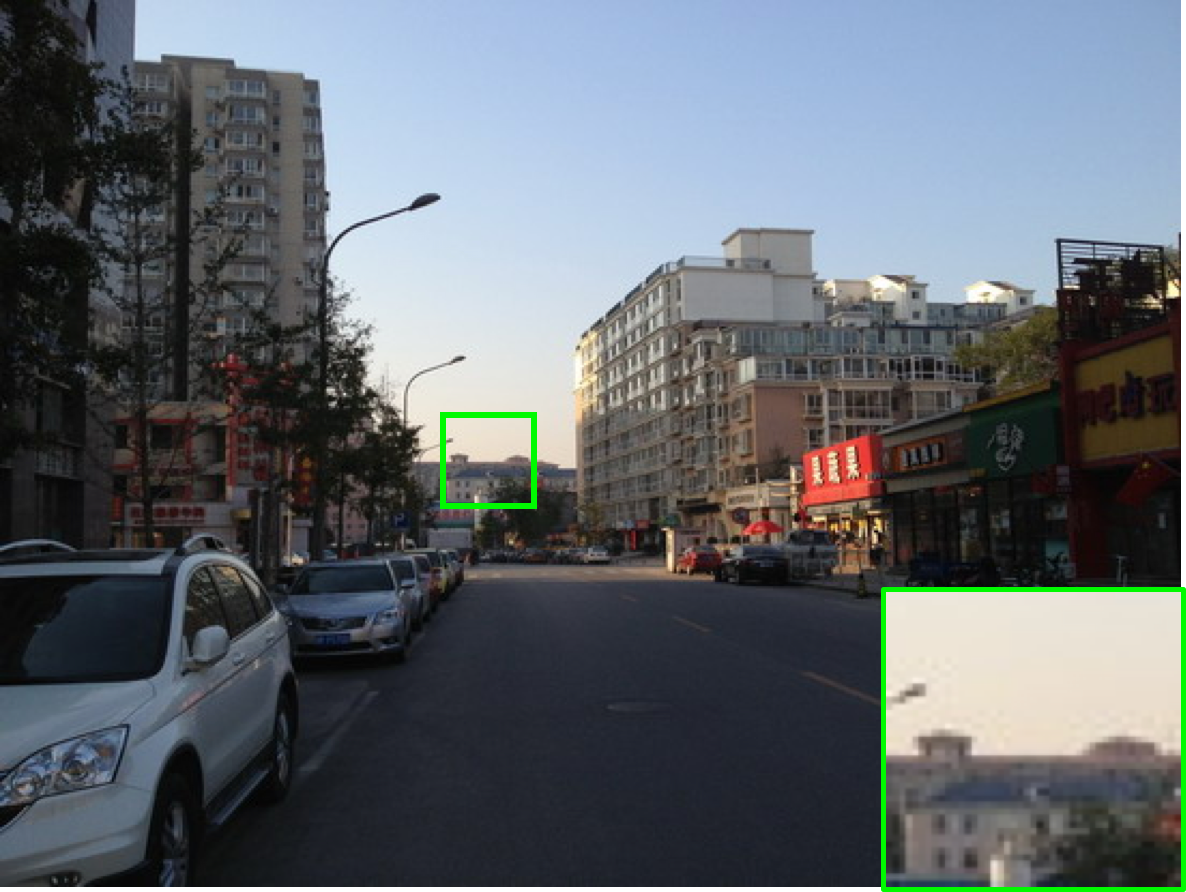}
		\scriptsize{(h) Ground truth}
	\end{minipage}
	\caption{Qualitative comparisons on SOTS.}
	\label{fig:SOTS}
\end{figure*}

\begin{figure*}[t]
	\centering
	\begin{minipage}[h]{0.13\linewidth}
		\centering
		\includegraphics[width=\linewidth]{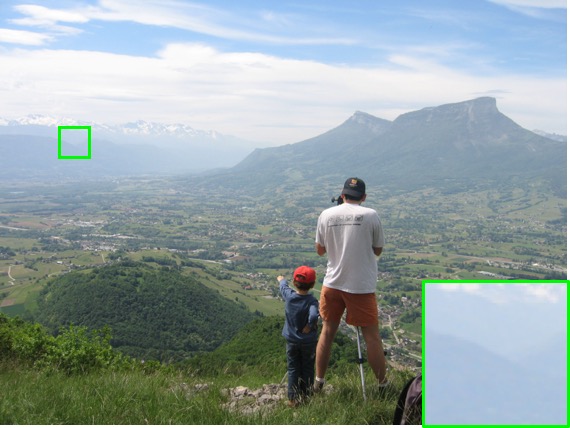}
	\end{minipage}
	\begin{minipage}[h]{0.13\linewidth}
		\centering
		\includegraphics[width=\linewidth]{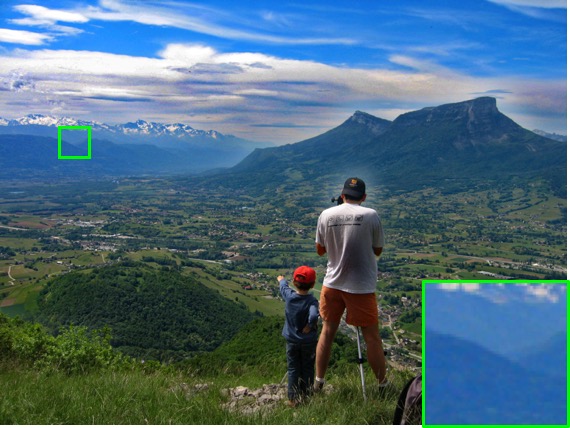}
	\end{minipage}
	\begin{minipage}[h]{0.13\linewidth}
		\centering
		\includegraphics[width=\linewidth]{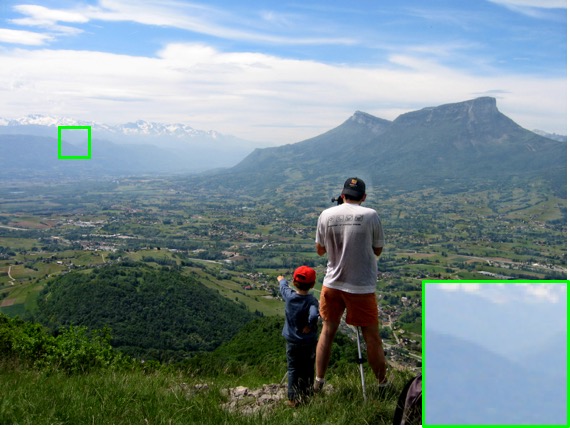}
	\end{minipage}
	\begin{minipage}[h]{0.13\linewidth}
		\centering
		\includegraphics[width=\linewidth]{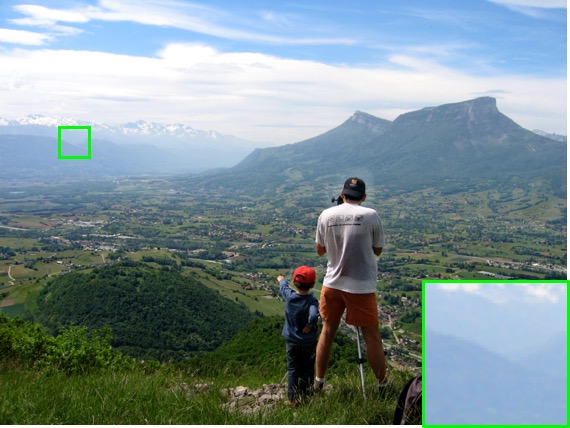}
	\end{minipage}
	\begin{minipage}[h]{0.13\linewidth}
		\centering
		\includegraphics[width=\linewidth]{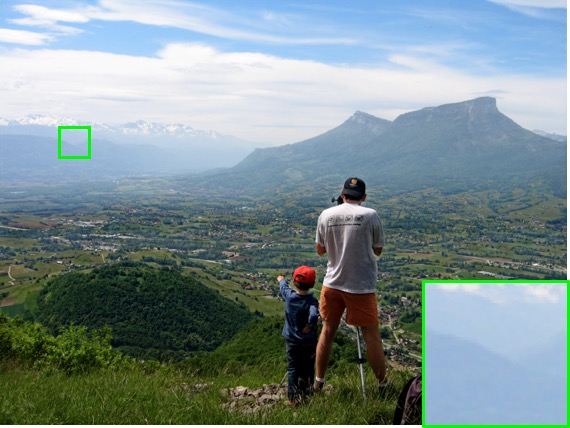}
	\end{minipage}
	\begin{minipage}[h]{0.13\linewidth}
		\centering
		\includegraphics[width=\linewidth]{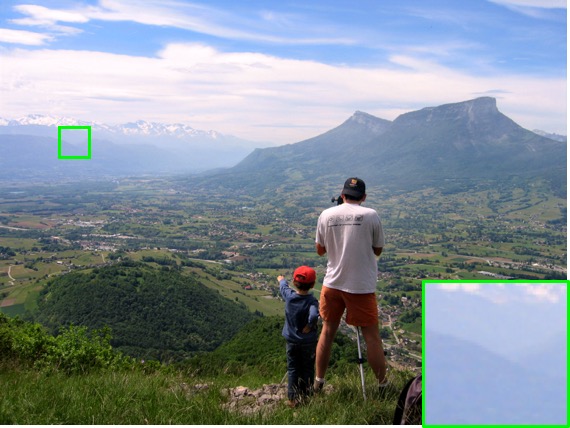}
	\end{minipage}
	\begin{minipage}[h]{0.13\linewidth}
		\centering
		\includegraphics[width=\linewidth]{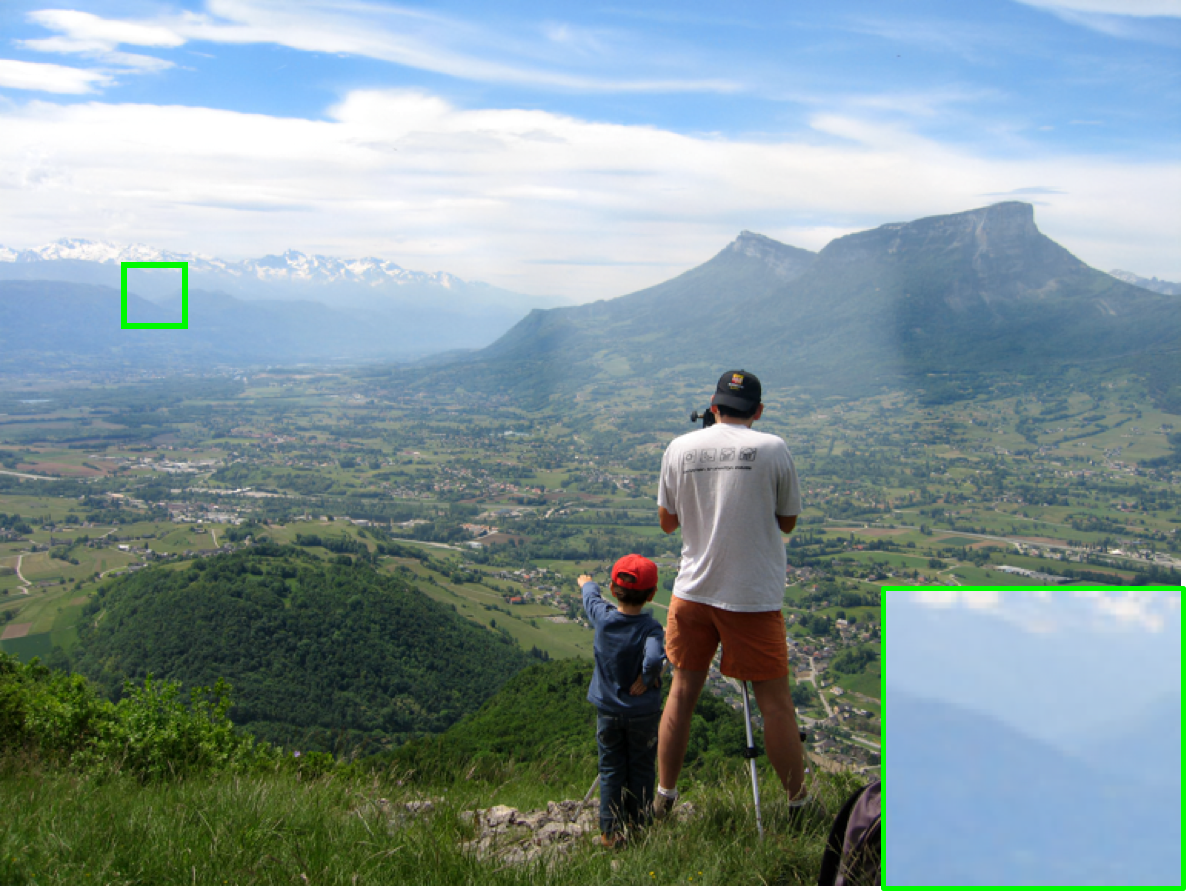}
	\end{minipage}
	\begin{minipage}[h]{0.13\linewidth}
		\includegraphics[width=\linewidth]{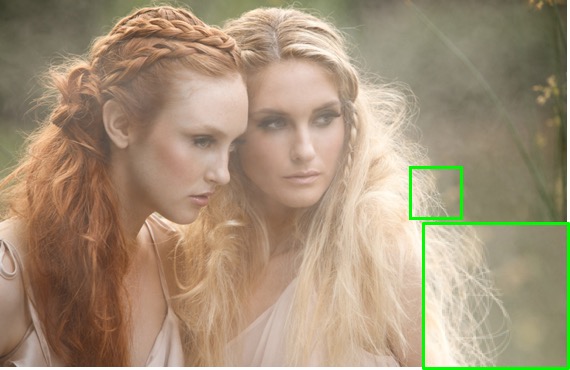}
	\end{minipage}
	\begin{minipage}[h]{0.13\linewidth}
		\centering
		\includegraphics[width=\linewidth]{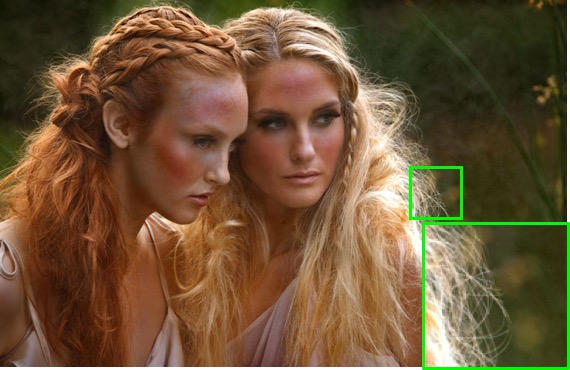}
	\end{minipage}
	\begin{minipage}[h]{0.13\linewidth}
		\centering
		\includegraphics[width=\linewidth]{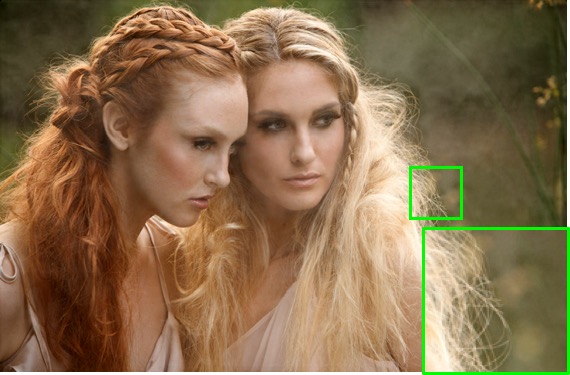}
	\end{minipage}
	\begin{minipage}[h]{0.13\linewidth}
		\centering
		\includegraphics[width=\linewidth]{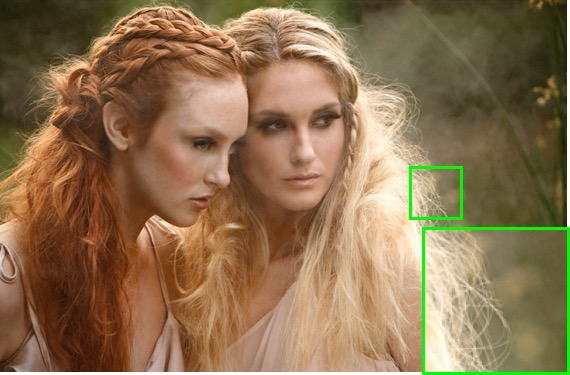}
	\end{minipage}
	\begin{minipage}[h]{0.13\linewidth}
		\centering
		\includegraphics[width=\linewidth]{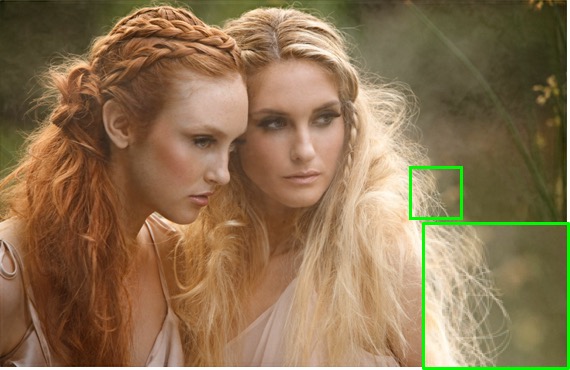}
	\end{minipage}
	\begin{minipage}[h]{0.13\linewidth}
		\centering
		\includegraphics[width=\linewidth]{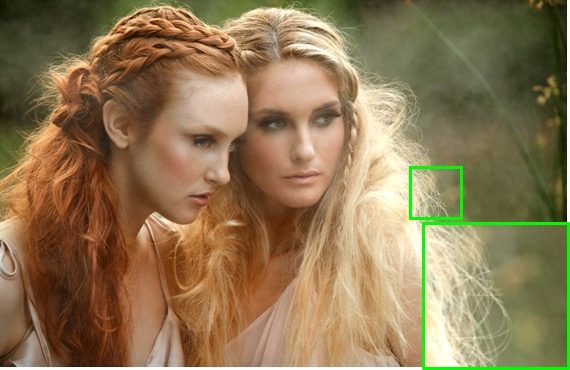}
	\end{minipage}
	\begin{minipage}[h]{0.13\linewidth}
		\centering
		\includegraphics[width=\linewidth]{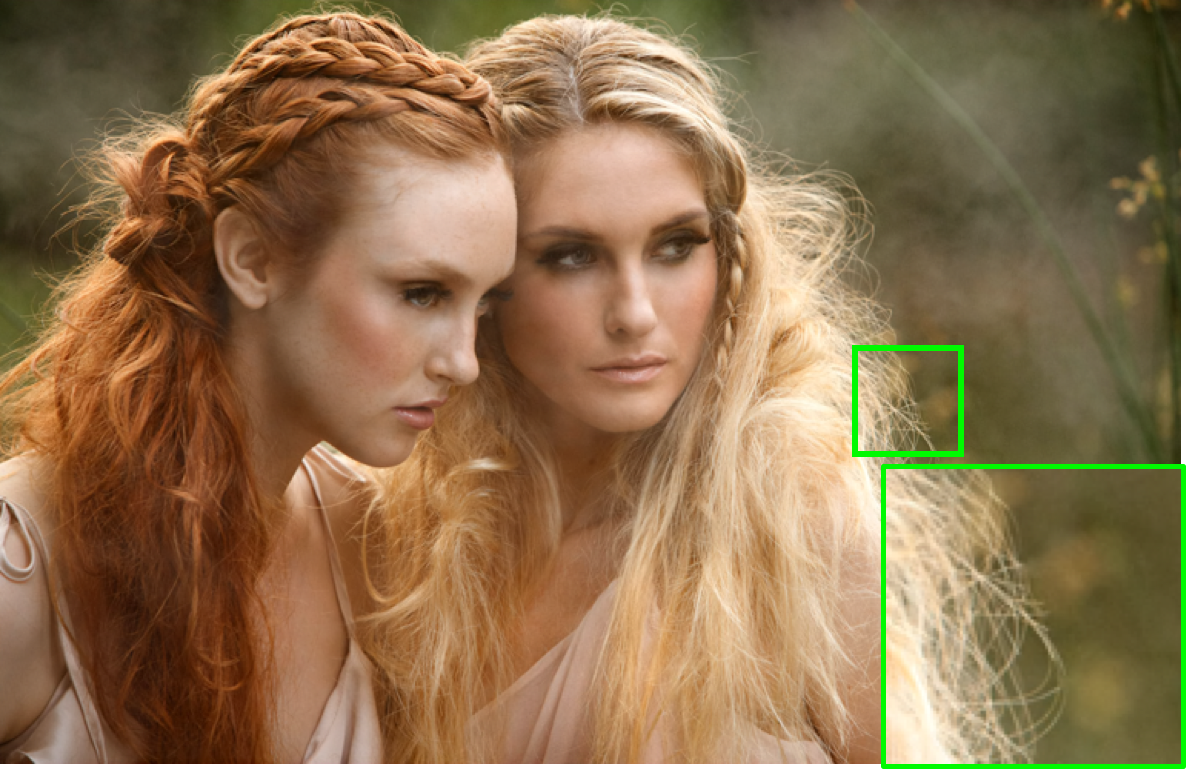}
	\end{minipage}
	\begin{minipage}[h]{0.13\linewidth}
		\centering
		\includegraphics[width=\linewidth]{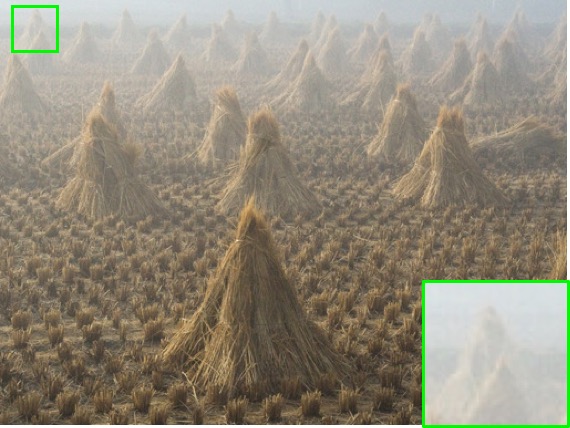}
	\end{minipage}
	\begin{minipage}[h]{0.13\linewidth}
		\centering
		\includegraphics[width=\linewidth]{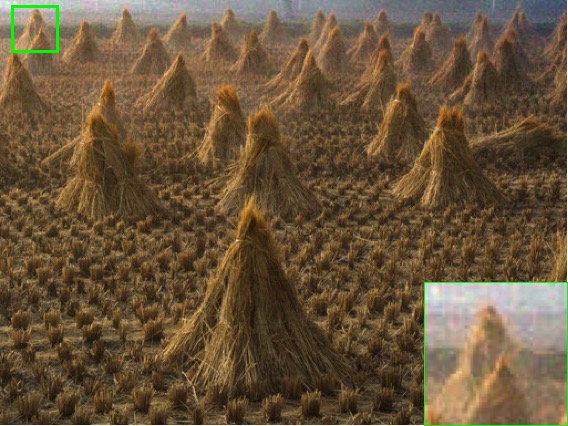}
	\end{minipage}
	\begin{minipage}[h]{0.13\linewidth}
		\centering
		\includegraphics[width=\linewidth]{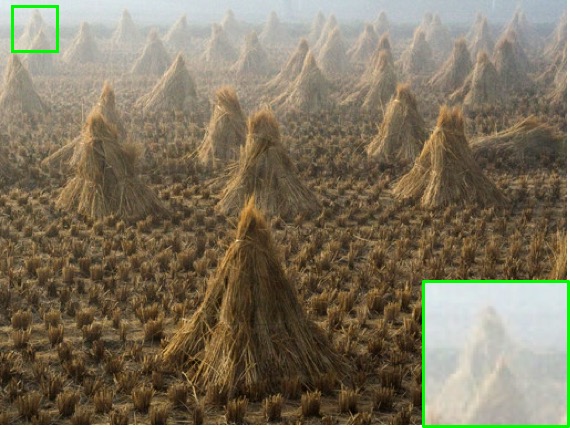}
	\end{minipage}
	\begin{minipage}[h]{0.13\linewidth}
		\centering
		\includegraphics[width=\linewidth]{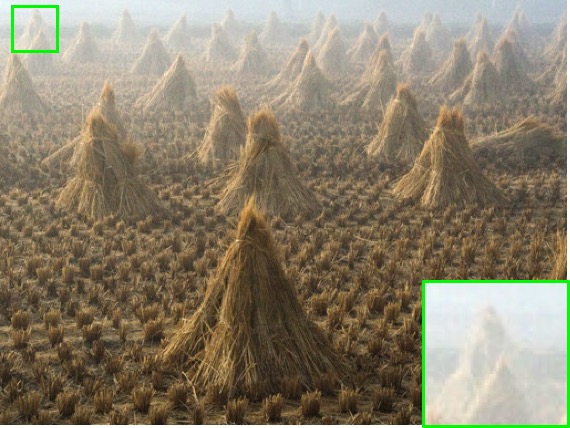}
	\end{minipage}
	\begin{minipage}[h]{0.13\linewidth}
		\centering
		\includegraphics[width=\linewidth]{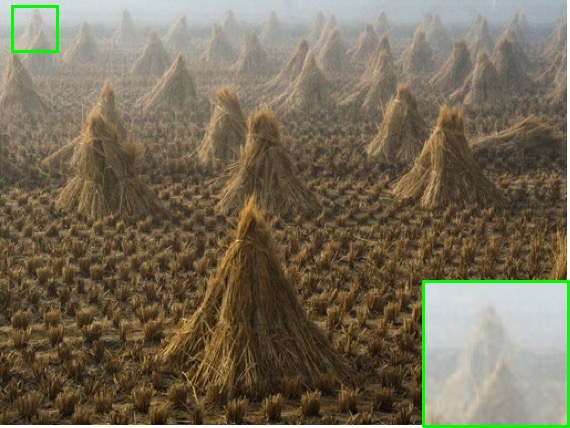}
	\end{minipage}
	\begin{minipage}[h]{0.13\linewidth}
		\centering
		\includegraphics[width=\linewidth]{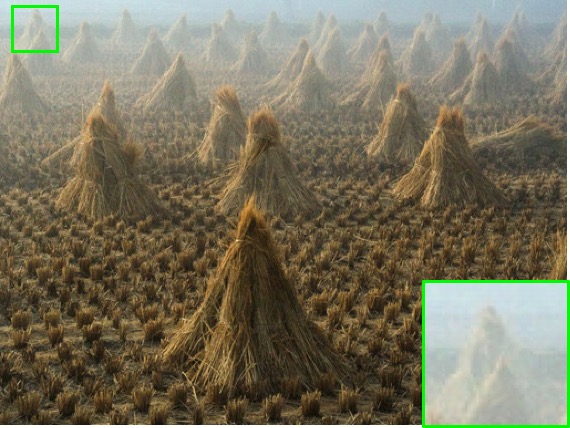}
	\end{minipage}
	\begin{minipage}[h]{0.13\linewidth}
		\centering
		\includegraphics[width=\linewidth]{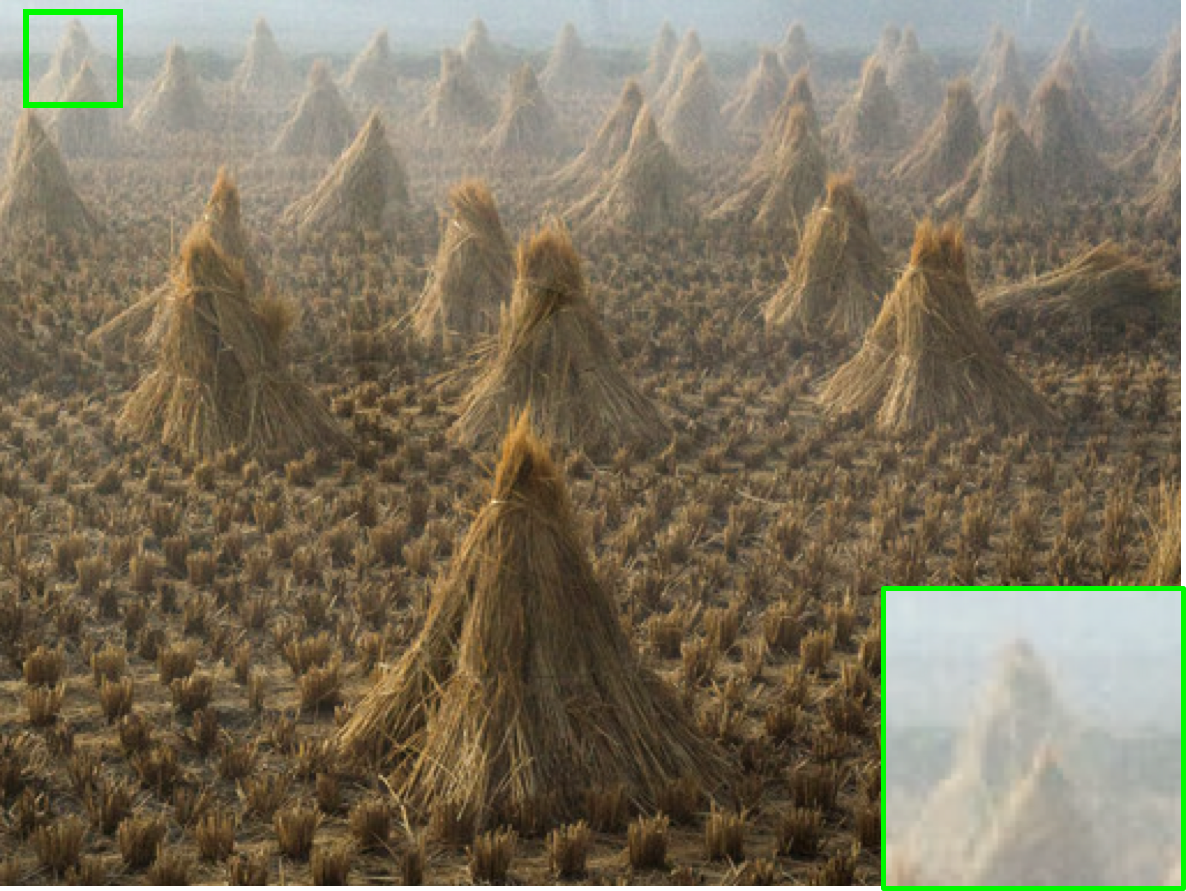}
	\end{minipage}
	\begin{minipage}[h]{0.13\linewidth}
		\centering
		\includegraphics[width=\linewidth]{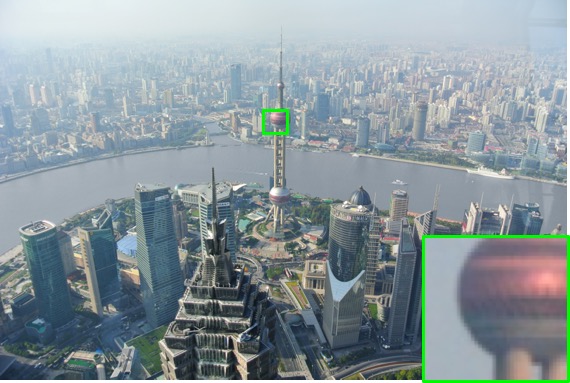}
		\scriptsize{(a) Hazy input}
	\end{minipage}
	\begin{minipage}[h]{0.13\linewidth}
		\centering
		\includegraphics[width=\linewidth]{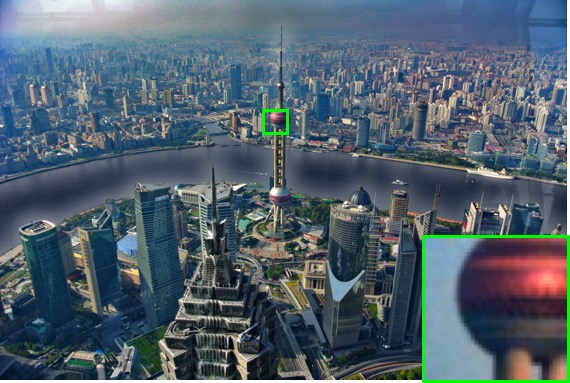}
		\scriptsize{(b) DCP}
	\end{minipage}
	\begin{minipage}[h]{0.13\linewidth}
		\centering
		\includegraphics[width=\linewidth]{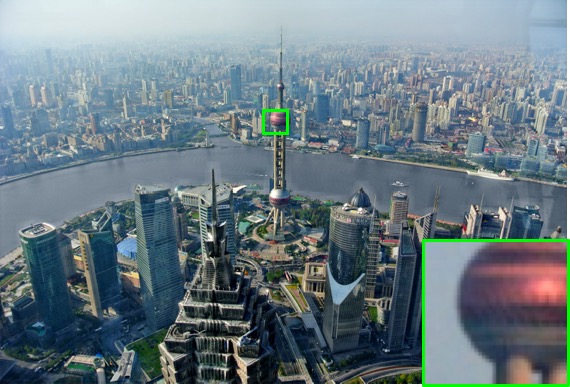}
		\scriptsize{(c) DehazeNet}
	\end{minipage}
	\begin{minipage}[h]{0.13\linewidth}
		\centering
		\includegraphics[width=\linewidth]{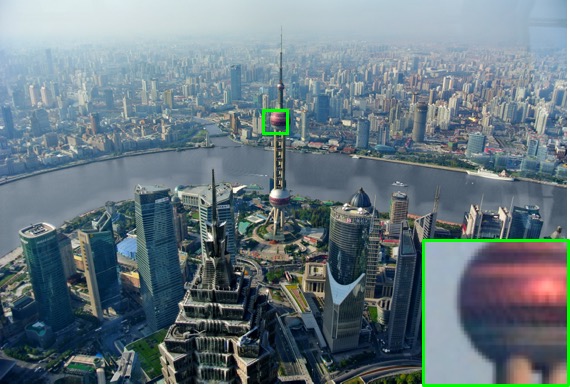}
		\scriptsize{(d) MSCNN}
	\end{minipage}
	\begin{minipage}[h]{0.13\linewidth}
		\centering
		\includegraphics[width=\linewidth]{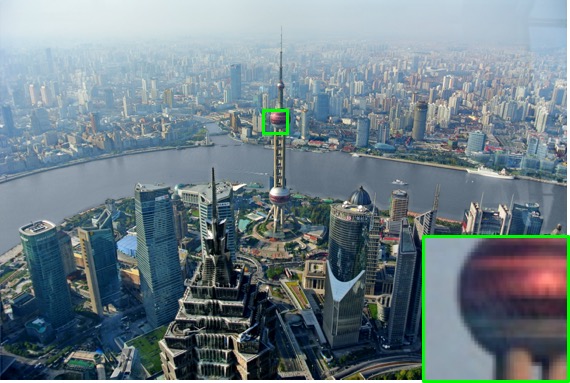}
		\scriptsize{(e) AOD-Net}
	\end{minipage}
	\begin{minipage}[h]{0.13\linewidth}
		\centering
		\includegraphics[width=\linewidth]{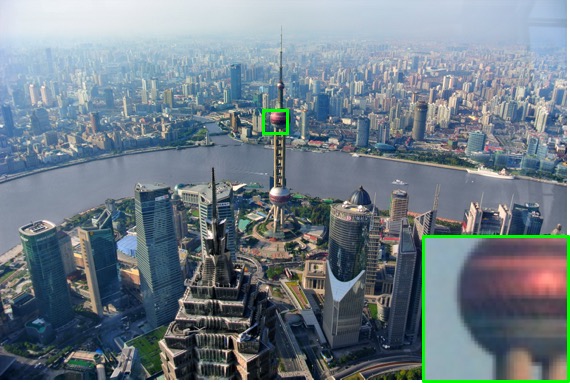}
		\scriptsize{(f) GFN}
	\end{minipage}
	\begin{minipage}[h]{0.13\linewidth}
		\centering
		\includegraphics[width=\linewidth]{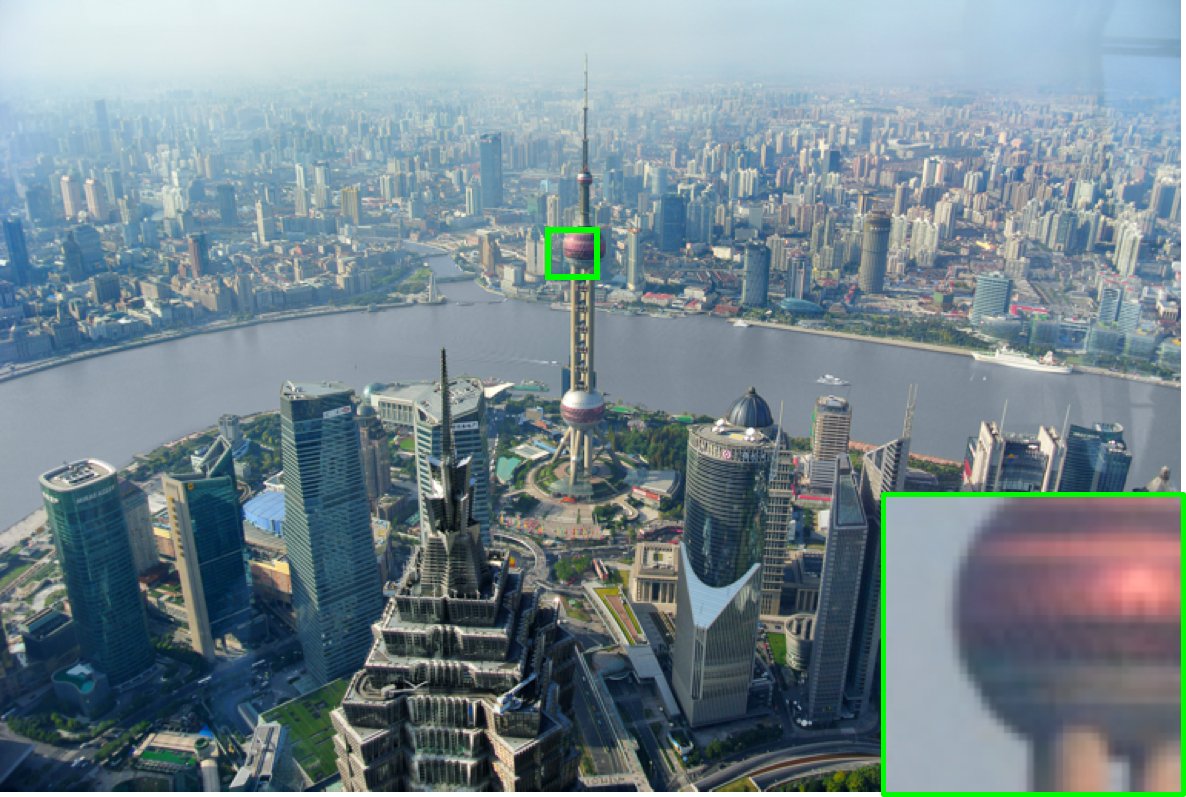}
		\scriptsize{(g) Ours}
	\end{minipage}
	\caption{Qualitative comparisons on the real-world dataset \cite{fattal2014dehazing}.}
	\label{fig:real}
\end{figure*}

\subsection{Synthetic Dataset}
The proposed network is tested on the synthetic dataset for qualitative and quantitative comparisons with the state-of-the-arts that include DCP~\cite{iebmdcp01}, DehazeNet~\cite{iebmdehazenet01}, MSCNN~\cite{iebmmscnn01}, AOD-Net~\cite{iebmaod01} and GFN~\cite{iebmgated01}. The DCP is a prior-based method and is regarded as the baseline in single image dehazing.
The others are data-driven methods. Moreover, except for AOD-Net and GFN, these methods all follow the same strategy of first estimating the transmission map and the atmosphere light then leveraging the atmosphere scattering model to compute the dehazed image. For fair comparisons, the above-mentioned data-driven methods are trained in the same way as the proposed one. The SOTS from RESIDE is employed as the testing dataset. We use peak signal to noise ratio (PSNR) and structure similarity (SSIM) for quantitative assessment of the dehazed outputs. 




Fig.~\ref{fig:SOTS} shows the qualitative comparisons on both synthetic indoor and outdoor images from SOTS. Due to the inaccurate estimation of haze thickness, the results of DCP are typically darker than the ground truth. Moreover, the DCP tends to cause severe color distortions, thereby jeopardizing the quality of its output (see, {\textit{e}.\textit{g}.}, the tree and the sky in Fig.~\ref{fig:SOTS} (b)). For DehazeNet as well as MSCNN,
a significant amount of haze still remains unremoved and the output suffers color distortions. The AOD-Net largely overcomes the color distortion problem, but it tends to cause  halo artifacts around object boundaries (see, {\textit{e}.\textit{g}., the chair leg in Fig.~\ref{fig:SOTS} (e))  and the removal of the hazy effect is visibly incomplete.	The GFN  succeeds in suppressing the halo artifacts to a certain extent.  However, it has limited ability to remove  thick haze (see, {\textit{e}.\textit{g}., the area between two chairs and the fireplace in Fig.~\ref{fig:SOTS} (f)). 
Compared with the state-of-the-arts, the proposed method has the best performance in terms of haze removal and artifact/distortion suppression (see, {\textit{e}.\textit{g}., Fig.~\ref{fig:SOTS} (g)). The dehazed images produced by GridDehazeNet are free of major artifacts/distortions and are visually most similar to their haze-free counterparts.
		 	
		 	
Table~\ref{tab:syn1} shows the quantitative comparisons on the SOTS in terms of average PSNR and SSIM values. We note that the proposed method outperforms the state-of-the-arts by a wide margin. 		 	
We have also tested these dehazing methods (all pre-trained on the
OTS dataset except for the DCP) directly on a new synthetic dataset. The
hazy images in this new dataset are generated from 500 clear images
(together with their depth maps) randomly selected
from the Sun RGB-D dataset~\cite{Song_2015_CVPR} through the atmosphere
scattering model with  $\beta\in[0.04,0.2]$ and $A\in[0.8,1.0]$. As shown in Table~\ref{tab:syn1}, the proposed method is fairly robust and continues to show highly competitive performance.

\begin{table}[h]
	\caption{\label{tab:syn1}Quantitative comparisons on SOTS and Sun RGB-D for different methods.}
	\vspace{-2mm}
	\footnotesize
	\center
	\begin{tabular}{|c|c|c|c|c|c|c|}
		\hline
		\multicolumn{1}{|c|}{\multirow{2}*{Method}}&\multicolumn{2}{|c|}{Indoor}&\multicolumn{2}{|c|}{Outdoor} &\multicolumn{2}{|c|}{Sun RGB-D}\\
		\cline{2-7}
		& PSNR & SSIM & PSNR & SSIM & PSNR & SSIM \\    %
		\hline	
		\hline
		DCP	&16.61	&0.8546	&19.14	&0.8605   &15.18   &0.8191\\
		\hline
		DehazeNet &19.82	&0.8209	&24.75	&0.9269 &23.05  &0.8870\\
		\hline
		MSCNN 	&19.84	&0.8327	&22.06	&0.9078 &23.85  &0.9095 \\
		\hline
		AOD-Net &20.51	&0.8162	&24.14	&0.9198 &22.51  &0.8918 \\
		\hline
		GFN	&24.91	&0.9186	&28.29	&0.9621 &25.35  &0.9250 \\
		\hline
		Ours	&\textbf{32.16}	&\textbf{0.9836}	&\textbf{30.86}	&\textbf{0.9819} & \textbf{28.67} & \textbf{0.9599}\\
		\hline
	\end{tabular}
	
\end{table}

\subsection{Real-World Dataset}

We further compare the proposed method against the state-of-the-arts on the real-world dataset~\cite{fattal2014dehazing}. Here we shall only make qualitative comparisons since the haze-free counterparts of the real-world hazy images in this dataset  are not available.
As shown by Fig \ref{fig:real}, the results   are largely consistent with those on the synthetic dataset. The DCP again suffers severe color distortions (see, \textit{e}.\textit{g}., the sky and the girls' face in Fig \ref{fig:real} (b)). For DehazeNet, MSCNN and AOD-Net, haze removal is clearly incomplete. The GFN has limited ability to deal with dense haze and causes color distortions in some cases (see, \textit{e}.\textit{g}., the sky and the piles in Fig \ref{fig:real} (f)). In comparison to the aforementioned methods, the proposed GridDehazeNet is more effective in haze removal and distortion suppression.
 



\subsection{Atmosphere Scattering Model}\label{sec:asm}

To gain a better understanding of the difference between the direct estimation strategy adopted by the proposed method (where the atmosphere scattering model is completely bypassed) and the indirect estimation strategy (where the transmission map and the atmospheric light intensity are first estimated, which are then leveraged to compute the dehazed image via the atmosphere scattering model), we repurpose the proposed GridDehazeNet for the estimation of the transmission map and the atmospheric light intensity. Specifically, we modify the convolutional layer at the output end ({\textit{i}.\textit{e}.,  the rightmost convolutional layer in Fig.~\ref{fig:GDN_main}) so that it outputs two feature maps, one as the estimated transmission map and the mean of the other as the estimated atmospheric light intensity; these two estimates are then substituted into Eq.~(\ref{eq:asm}) to determine the dehazed image. The resulting network is trained in the same way as before and is tested on both SOTS and Sun RGB-D. Although adopting the atmosphere scattering model leads to a significant reduction in the number of parameters that need to be estimated, it in fact incurs performance degradation as shown in Table~\ref{tab:estimation}. 
This indicates that incorporating the atmosphere scattering model into the proposed network does have a detrimental effect on the loss surface.

\begin{table}[h]
	\caption{\label{tab:estimation}Comparisons for different estimation strategies. }
	\vspace{-2mm}
	\footnotesize
	\center
	\begin{tabular}{|c|c|c|c|c|c|c|}
		\hline
		\multicolumn{1}{|c|}{\multirow{2}*{Estimation}} &\multicolumn{2}{|c|}{Indoor} &\multicolumn{2}{|c|}{Outdoor} &\multicolumn{2}{|c|}{SUN RGB-D}\\
		\cline{2-7}
		& PSNR & SSIM & PSNR & SSIM & PSNR & SSIM \\    %
		\hline	
		\hline
		Indirect & 30.33&	0.9160&	 30.12&	 0.9729&  27.82&  0.9477\\
		\hline
		Direct	&\textbf{32.16}	&\textbf{0.9836}	&\textbf{30.86}	&\textbf{0.9819} &\textbf{28.67} &\textbf{0.9599}\\
		\hline
	\end{tabular}
	\vspace{-3mm}
\end{table}

\subsection{Learned Inputs}

Fig.~\ref{fig:CNN_priors} illustrates four learned inputs (out of a total of 16 learned inputs) generated by the pre-processing module. It can be seen that each learned input enhances a certain aspect of the given hazy image. For instance, the learned input with index 9 highlights a specific texture, which is not evidently shown in the hazy image.

\begin{figure}[h]
	\centering
	\begin{minipage}[h]{0.32\linewidth}
		\centering
		\includegraphics[width=\linewidth]{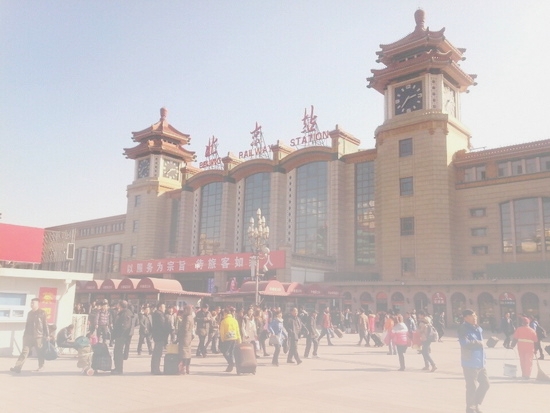}
		\scriptsize{(a) Hazy image}
	\end{minipage}
	\begin{minipage}[h]{0.32\linewidth}
		\centering
		\includegraphics[width=\linewidth]{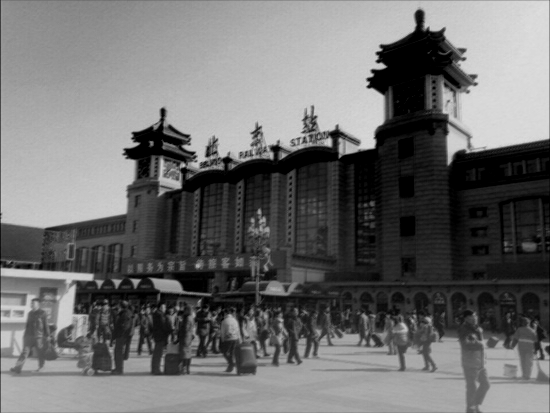}
		\scriptsize{(c) Learned input (index 0)}
	\end{minipage}
	\begin{minipage}[h]{0.32\linewidth}
		\centering
		\includegraphics[width=\linewidth]{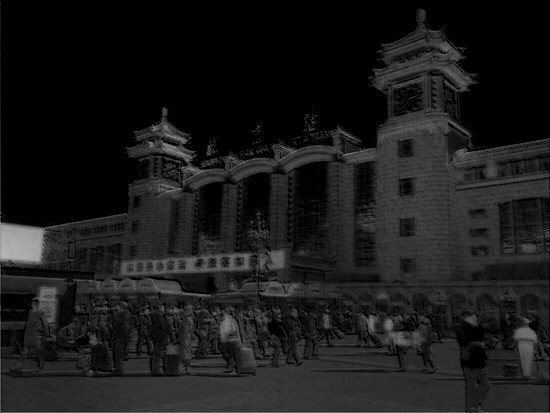}
		\scriptsize{(e) Learned input (index 8)}
	\end{minipage}
	\begin{minipage}[h]{0.32\linewidth}
		\centering
		\includegraphics[width=\linewidth]{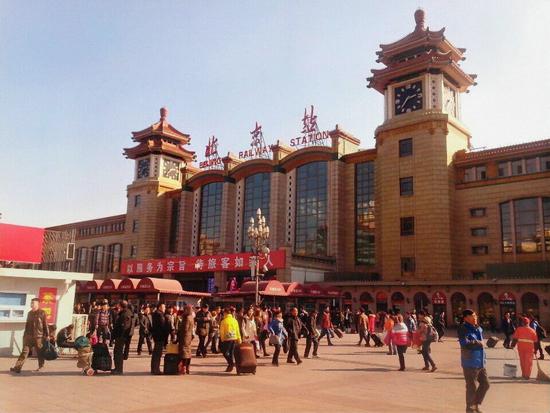}
		\scriptsize{(b) Dehazed image}
	\end{minipage}
	\begin{minipage}[h]{0.32\linewidth}
		\centering
		\includegraphics[width=\linewidth]{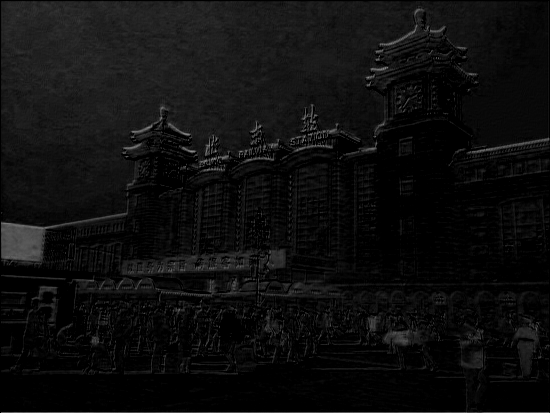}
		\scriptsize{(d) Learned input (index 1)}
	\end{minipage}
	\begin{minipage}[h]{0.32\linewidth}
		\centering
		\includegraphics[width=\linewidth]{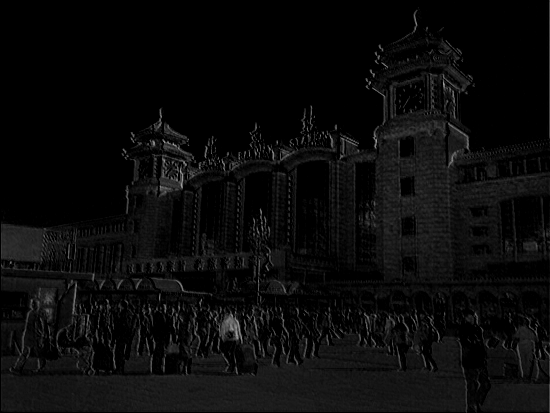}
		\scriptsize{(f) Learned input (index 9)}
	\end{minipage}
	\caption{Visualization of the hazy image, the dehazed image and several learned inputs.}
	\label{fig:CNN_priors}
\end{figure}

We conduct the following experiment to demonstrate the diversity gain offered by the learned inputs. Specifically, we remove the pre-processing module and
replace the first three learned inputs by the RGB channels of the given hazy image and the rest by all-zero feature maps. We also conduct an experiment to show the advantages of learned inputs over those derived inputs produced by hand-selected pre-processing methods. 
In this case, we replace the learned inputs by the same number of derived inputs (three from the given hazy image, three from the white balanced (WB) image, three from the contrast enhanced (CE) image, three from the gamma corrected (GC) image, three from the gamma corrected GC image and one from the gray scale image). Here the use of WB, CE, GC images as derived inputs is inspired by~\cite{iebmgated01}. In both cases, the resulting networks are trained in the same way as before and are tested on the SOTS. As shown in
Table~\ref{tab:input}, the learned inputs offer significant diversity gain and have clear advantages over the derived inputs.

\begin{table}[h]
	\caption{\label{tab:input}Comparisons on SOTS for different types of inputs. }
	\vspace{-2mm}
	\footnotesize
	\center
	\begin{tabular}{|c|c|c|c|c|}
		\hline
		\multicolumn{1}{|c|}{\multirow{2}*{Input}}&\multicolumn{2}{|c|}{Indoor}&\multicolumn{2}{|c|}{Outdoor}\\
		\cline{2-5}
		& PSNR & SSIM & PSNR & SSIM \\    %
		\hline	
		\hline
		Original 	&31.48	&0.9820	&30.33	&0.9808\\
		\hline
		Derived 	&30.21	&0.9799	&30.32	&0.9778\\
		\hline
		Learned	&\textbf{32.16}	&\textbf{0.9836}	&\textbf{30.86}	&\textbf{0.9819}\\
		\hline
	\end{tabular}
	\vspace{-3mm}
	
\end{table}

\subsection{Ablation Study}

We perform ablation studies by considering different configurations of the backbone module of the proposed GridDehazeNet. 
Note that each row in the backbone module corresponds to a different scale, and the columns in the backbone module serve as bridges to facilitate the information exchange across different scales. 
Table~\ref{tab:syn2} shows how the performance of the proposed GridDehazeNet depends on 
the number of rows (denoted by $r$) and the number of columns (denoted by $c$) in the backbone module.
It is clear that increasing $r$ and $c$ leads to higher average PSNR and SSIM values.



\begin{table}[h]
	\caption{\label{tab:syn2}Comparisons on SOTS for  different configurations. }
	\footnotesize
	\vspace{-2mm}
	\center
	\begin{tabular}{| c| c|c|c|c|c|}
		\hline
		\multicolumn{2}{|c|}{\multirow{2}*{Configuration}} & \multicolumn{2}{|c|}{Indoor} & \multicolumn{2}{|c|}{Outdoor}\\
		\cline{3-6}
		\multicolumn{2}{|c|}{}  &   PSNR    &SSIM   &PSNR   &SSIM\\
		\hline
		\hline
		\multirow{3}*{$r=1$}	&	$c=2$	&22.38	&0.8849 &25.64  &0.9435	\\
		&	$c=4$	&24.92	&0.9375	&27.32   &0.9619\\
		&	$c=6$	&25.95	&0.9507	&27.84   &0.9676\\
		\cline{1-6}
		\multirow{3}*{$r=2$}	&	$c=2$	&22.53	&0.8931 &25.71   &0.9444	\\
		&	$c=4$	&26.96	&0.9581	&28.47   &0.9716\\
		&	$c=6$	&28.64	&0.9701	&29.12   &0.9760\\
		\cline{1-6}
		\multirow{3}*{$r=3$}	&	$c=2$	&22.57	&0.8951	&25.73  &0.9439\\
		&	$c=4$	&29.40	&0.9752	&29.96  &0.9795\\
		&	$c=6$	&\textbf{32.16}	&\textbf{0.9836}	&\textbf{30.86}   &\textbf{0.9819}	\\
		\hline
	\end{tabular}
	\vspace{-3mm}
\end{table}

We perform further ablation studies by considering several variants of the proposed GridDehazeNet, which include the original GridNet~\cite{iebmgridnet01}, the multi-scale network resulted from removing the exchange branches (except for the first and the last ones that are needed to  maintain the minimum connection), our model without attention-based channel-wise feature fusion, without the post-processing module or without perceptual loss, as well as the encoder-decoder network obtained by pruning the proposed network (see the red path in Fig.~\ref{fig:GDN_main} ). These variants are all trained in the same way as before and are tested on the SOTS. As shown in Table~\ref{tab:syn3}, each component has its own contribution to the performance of the full model, which justifies the overall design. 

\begin{table}[h]
	\caption{\label{tab:syn3}Comparisons on SOTS for different variants of GridDehazeNet.}
	\vspace{-2mm}
	\footnotesize
	\center
	\begin{tabular}{|c|c|c|c|c|}
		\hline
		\multirow{2}*{Variant}  &   \multicolumn{2}{|c|}{Indoor}    &   \multicolumn{2}{|c|}{Outdoor}\\
		\cline{2-5}
		&   PSNR    &SSIM   &PSNR   &SSIM\\
		\hline
		\hline
		Original GridNet~\cite{iebmgridnet01}    &27.37   & 0.9267  &28.30  &0.9307\\
		\hline
		w/o exchange branches &29.57   &0.9765   &30.18   &0.9795\\
		\hline
		w/o attention   &31.77   &0.9833  &30.32  &0.9809\\
			\hline
		w/o post-processing   &31.62   &0.9779   &30.52   &0.9810\\
		\hline
		w/o perceptual loss &31.83  &0.9815   &30.51  &0.9768\\
		\hline
	encoder-decoder &28.48   &0.9662   &28.61   &0.9715\\
		\hline
		Our full model  &\textbf{32.16}   &\textbf{0.9836}   &\textbf{30.86}   &\textbf{0.9819}\\
		\hline
	\end{tabular}
	\vspace{-3mm}
	
\end{table}

\subsection{Runtime Analysis}
Our un-optimized code takes about 0.22$s$ to dehaze one image from SOTS on average. We have also evaluated the computational efficiency of the aforementioned state-of-the-art methods and plot their average runtimes in Fig.~\ref{fig:CC}. It can be seen that the proposed GridDehazeNet ranks second among the dehazing methods under comparison. 
\begin{figure}[h]
	\vspace{-3mm}
	\centering
	\includegraphics[width=\linewidth]{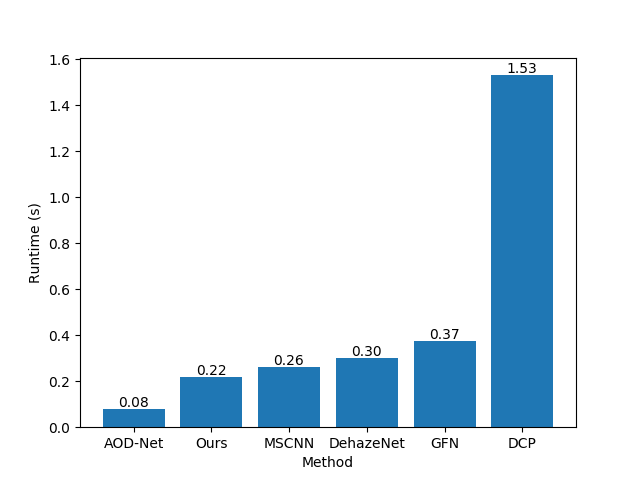}
	\caption{Runtime comparison of different dehazing methods.}
	\label{fig:CC}
\end{figure}

\section{Conclusion}
We have proposed an end-to-end trainable CNN, named GridDehazeNet, and demonstrated its competitive performance for single image dehazing. Due to the generic nature of its building components, the proposed GridDehazeNet is expected to be applicable to a wide range of image restoration problems. Our work also sheds some light on the puzzling phenomenon concerning the use of the atmosphere scattering model in image dehazing, and suggests the need to rethink the role of physical model in the design of image restoration algorithms.


{\small
	\bibliographystyle{ieee_fullname}
	\bibliography{egbib}
}

\end{document}